\documentclass[10pt,journal,compsoc]{IEEEtran}

\ifCLASSOPTIONcompsoc
  \usepackage[nocompress]{cite}
  \usepackage[caption=false,font=footnotesize,labelfont=sf,textfont=sf]{subfig}
\else
  \usepackage{cite}
  \usepackage[caption=false,font=footnotesize]{subfig}
\fi
\ifCLASSINFOpdf
  \usepackage[pdftex]{graphicx}
  \DeclareGraphicsExtensions{.pdf,.jpeg,.png}
\else
  \usepackage[dvips]{graphicx}
  \DeclareGraphicsExtensions{.eps}
\fi
\usepackage{amsmath,amssymb,amsfonts,dsfont,pifont,bm,bbm,mathrsfs,mathtools}
\usepackage{algorithm,algpseudocode,listings}
\usepackage{booktabs,multirow,adjustbox,diagbox}
\usepackage{xspace}
\usepackage{float,capt-of}
\usepackage[table]{xcolor}
\usepackage{url}
\usepackage{ulem}
\usepackage[pagebackref=true,breaklinks=true,colorlinks=true,citecolor=blue,bookmarks=false]{hyperref}
\usepackage[capitalize]{cleveref}  
\crefname{section}{Sec.}{Secs.}
\Crefname{section}{Section}{Sections}
\crefname{table}{Tab.}{Tabs.}
\Crefname{table}{Table}{Tables}
\crefname{figure}{Fig.}{Figs.}
\Crefname{figure}{Figure}{Figures}
\crefname{equation}{Eq.}{Eqs.}
\Crefname{equation}{Equation}{Equations}
\newcommand{\blocks}[3]{\multirow{3}{*}{\(\left[\begin{array}{c}\text{1$\times$1, #2}\\[-.1em] \text{3$\times$3, #2}\\[-.1em] \text{1$\times$1, #1}\end{array}\right]\)$\times$#3}
}

\newcommand{\z}{{\rm\bf z}}         
\newcommand{\Z}{\mathcal{Z}}        
\newcommand{\w}{{\rm\bf w}}         
\newcommand{\W}{\mathcal{W}}        
\newcommand{\y}{{\rm\bf y}}         
\newcommand{\Y}{\mathcal{Y}}        
\newcommand{\x}{{\rm\bf x}}         
\newcommand{\f}{{\rm\bf f}}         
\newcommand{\E}{\mathbb{E}}         
\newcommand{\FID}{FID$\downarrow$}  
\newcommand{\SSIM}{SSIM$\uparrow$}  
\newcommand{\MSE}{MSE$\downarrow$}  
\newcommand{\TIME}{TIME$\downarrow$}  
\newcommand{\method}{GH-Feat\xspace}

\definecolor{revision_color}{rgb}{1.0, 0.5, 0.0}

\newcommand{\revise}[1]{#1}

\begin{document}

\newcommand{\titlename}{GH-Feat: Learning Versatile Generative Hierarchical Features from GANs}
\title{\titlename}

\author{
  Yinghao Xu,
  Yujun Shen,
  Jiapeng Zhu,
  Ceyuan Yang, and
  Bolei Zhou, \IEEEmembership{Member, IEEE}%
  \IEEEcompsocitemizethanks{
    \IEEEcompsocthanksitem Y. Xu and C. Yang are with
    the Department of Information Engineering, the Chinese University of Hong Kong, Hong Kong SAR.\protect
    \IEEEcompsocthanksitem Y. Shen is with
    Ant Research, China.\protect
    \IEEEcompsocthanksitem J. Zhu is with
    the School of Computer Science and Engineering, the Hong Kong University of Science and Technology, Hong Kong SAR.\protect
    \IEEEcompsocthanksitem B. Zhou is with Computer Science Department, University of California, Los Angeles, USA.
  }%
}


\IEEEtitleabstractindextext{
    \begin{abstract}

Recent years witness the tremendous success of generative adversarial networks (GANs) in synthesizing photo-realistic images.
GAN generator learns to compose realistic images and reproduce the real data distribution. Through that, a hierarchical visual feature with multi-level semantics spontaneously emerges.
In this work we investigate that such a generative feature learned from image synthesis exhibits great potentials in solving a wide range of computer vision tasks, including both generative ones and more importantly discriminative ones.
We first train an encoder by considering the pre-trained StyleGAN generator as a learned loss function.
The visual features produced by our encoder, termed as \textit{Generative Hierarchical Features (\method)}, highly align with the layer-wise GAN representations, and hence describe the input image adequately from the reconstruction perspective.
Extensive experiments support the versatile transferability of \method across a range of applications, such as image editing, image processing, image harmonization, face verification, landmark detection, layout prediction, image retrieval, \textit{etc.}
We further show that, through a proper spatial expansion, our developed \method can also facilitate fine-grained semantic segmentation using only a few annotations.
Both qualitative and quantitative results demonstrate the appealing performance of \method.
Code and models are available at  \protect\url{https://genforce.github.io/ghfeat/}.

\end{abstract}

}

\maketitle
\IEEEdisplaynontitleabstractindextext
\IEEEpeerreviewmaketitle

\IEEEraisesectionheading{\section{Introduction}\label{sec:introduction}}

\IEEEPARstart{R}{epresentation} learning plays an essential role in the rise of deep learning.
The learned representations are able to express the variation factors of the complex visual world.
Accordingly, the performance of a deep learning algorithm largely depends on the features extracted from the input data.
As pointed out by Bengio~\textit{et al.}~\cite{bengio2013representation}, a good representation is expected to have the following properties.
First, it should capture multiple configurations of the input.
Second, it should organize the explanatory factors of the input data as a hierarchy, where more abstract concepts are at a higher level.
Third, it should have strong transferability, not only from datasets to datasets but also from tasks to tasks.

Deep neural networks supervisedly trained for image classification on large-scale datasets (\textit{e.g.}, ImageNet~\cite{imagenet} and Places~\cite{zhou2017places}) have resulted in expressive and discriminative visual features~\cite{sharif2014cnn}.
However, the developed features are heavily dependent on the training objective.
For example, the deep features learned for the object recognition task may give more attention to the shapes and parts of the objects while remain invariant to rotation~\cite{bau2017network,matthew2014visualizing}, and the deep features from a scene classification model may focus more on detecting the categorical objects (\textit{e.g.}, bed for bedroom and sofa for living room)~\cite{zhou2015object}.
Thus the discriminative features learned from solving high-level image classification tasks might not be necessarily good for other mid-level and low-level tasks, limiting their transferability~\cite{yosinski2014transferable,zhao2020makes}.
Besides, it remains unknown how the discriminative features can be used in generative applications like image editing.

In contrast to discriminative models, generative models such as generative adversarial networks (GANs)~\cite{gan} offer an alternative way for representation learning.
%
%
A GAN is typically formulated to match the synthetic data distribution to the observed data distribution. 
Through adversarial training, the generator in GAN is required to capture the multi-level variation factors underlying the input data to the most extent, otherwise, the discrepancy between the real and synthesized data would be spotted by the discriminator.
Recent studies have confirmed that the StyleGAN family~\cite{stylegan,stylegan2} spontaneously encodes rich semantics in a hierarchical manner~\cite{interfacegan,higan}.
But the transferability of the per-layer representation learned by GANs is not fully verified in the literature.
Some attempts have been made to apply the generative representations (\textit{i.e.}, the representations emerging from solving generative tasks) to the high-level image classification task~\cite{bigan,ali,bigbigan}, yet leaving mid-level and low-level tasks less explored.

In this work, we make a thorough investigation into the utility of GAN representations and demonstrate their wide applications to downstream tasks, including both generative ones and more importantly discriminative ones.
To appropriately obtain the GAN-derived features from a given image, we train a hierarchical encoder using the pre-trained StyleGAN generator as a learned loss function.
Through that way, our encoder works together with the generator to reconstruct the image, implicitly required to extract the variation factors that describe the input.
We call the output visual features as \textit{Generative Hierarchical Features (\method)}.
We also observe in the encoder training that, only exploiting the supervision at the image level (\textit{i.e.}, the per-pixel reconstruction loss) may cause the overfitting of pixel values, severely limiting the transferability of the extracted representations.
To mitigate such a negative effect, we introduce a training regularizer from the statistical perspective. It alleviates the problem of distribution mismatching between the learned \method and the native GAN representations.

After the encoder is well prepared, we evaluate the resulting \method on a broad range of downstream applications.
On one hand, we verify that the representations learned by GANs naturally support various generative tasks.
Concretely, to edit or process target images, we can simply modulate their corresponding features, and reuse the generator (\textit{i.e.}, the same as the one used in encoder training) as a renderer to decode features back to images.
The extensive experiments show that our approach achieves global and local image manipulation, transferring styles between two images, image colorization, image inpainting, image super-resolution, \textit{etc.}
Besides, our \method also allows fusing objects from an image into another as the application of image harmonization.
On the other hand, we are interested in the discriminative capability of the generative features.
For this purpose, we treat the features extracted by our encoder as the base representations, on top of which we learn different linear task heads for a range of high-level and middle-level tasks.
Experiments on multiple datasets validate the effectiveness of \method on large-scale image classification, face verification, facial landmark detection, room layout prediction, transferring learning, image retrieval, \textit{etc.}
Furthermore, to enable dense prediction tasks, we manage to expand \method along spatial dimensions with an adequate modification of our encoder.
The improved spatial-aware visual features suggest compelling performance on fine-grained semantic segmentation using only a few annotations.

The preliminary result of this work is published at \cite{xu2021generative} as oral presentation.
We include the following new contents as the extension to the conference paper.
(1) We find the limitation of only using image-level supervision for encoder training, and provide an effective solution by introducing a distribution-level regularizer. Analyses and improvements are illustrated in \cref{exp:reg}.
(2) We include two more image editing tasks, \textit{i.e.} style transfer (\cref{exp:styletransfer}) and semantic manipulation (\cref{exp:semanticediting}), to validate that our \method can describe images moderately, aligning with human perception.
(3) We confirm in \cref{exp:image-processing} that \method also facilitates conventional image processing tasks, including image colorization, image inpainting, and image super-resolution.
(4) We include image retrieval as an addition discriminative task to verify the hierarchical property of \method, whose details are explained in \cref{exp:retrieval}.
(5) We propose a spatial expansion to our \method via learning a spatial-aware encoder, and show the great potential of the improved representations in data-efficient fine-grained semantic segmentation in \cref{exp:spatial-task}.

\section{Related Work}\label{sec:related-work}

\noindent\textbf{Visual Features.}
Visual Feature plays a fundamental role in the computer vision field.
Traditional methods used manually designed features~\cite{sift,surf,hog} for pattern matching and object detection.
These features are significantly improved by deep models~\cite{alexnet,vgg,resnet}, which automatically learn the feature extraction from large-scale datasets.
However, the features supervisedly learned for a particular task could be biased to the training task and hence become difficult to transfer to other tasks, especially when the target task is too far away from the base task~\cite{yosinski2014transferable,zhao2020makes}.
Unsupervised representation learning is widely explored to learn a more general and transferable feature~\cite{doersch2017multi,zhang2017split,wu2018unsupervised,gidaris2018unsupervised,hjelm2019learning,zhuang2019local,moco,cpc,cpc2,cmc}.
However, most of existing unsupervised feature learning methods focus on evaluating their features on the tasks of image recognition, yet seldom evaluate them on other mid-level or low-level tasks, let alone generative tasks.
Shocher \textit{et al.}~\cite{shocher2020semantic} discover the potential of discriminative features in image generation, but the transferability of such features are not fully verified.

\noindent\textbf{Generative Adversarial Networks.}
GANs~\cite{gan} are able to produce photo-realistic images via learning the underlying data distribution.
The recent advance of GANs~\cite{dcgan,pggan,biggan} has significantly improved the synthesis quality.
StyleGAN~\cite{stylegan} proposes a style-based generator with multi-level style codes and achieves the start-of-the-art generation performance.
However, little work explores the representation learned by GANs as well as how to apply such representation for other applications.
Some recent work interprets the semantics encoded in the internal representation of GANs and applies them for image editing~\cite{gansteerability,interfacegan,gandissect,mganprior,higan,idinvert}.
But it remains much less explored whether the learned GAN representations are transferable to discriminative tasks.

\noindent\textbf{Adversarial Representation Learning.}
The main reason of hindering GANs from being applied to discriminative tasks comes from the lack of inference ability.
To fill this gap, prior work introduces an additional encoder to the GAN structure~\cite{bigan,ali}.
Donahue and Simonyan~\cite{bigbigan} and Pidhorskyi \textit{et al.}~\cite{alae} extend this idea to the state-of-the-art BigGAN~\cite{biggan} and StyleGAN~\cite{stylegan} models respectively.
In this paper, we also study the representation learning using GANs, with following \textbf{improvements} compared to existing methods.
First, we propose to treat the well-trained StyleGAN generator as \textit{a learned loss function}.
Second, instead of mapping the images to the initial GAN latent space, like most algorithms~\cite{bigan,ali,bigbigan,alae} have done, we design a novel encoder to produce \textit{hierarchical} features that well align with the layer-wise representation learned by StyleGAN.
Third, besides the image classification task that is mainly targeted at by prior work~\cite{bigan,ali,bigbigan,alae}, we validate the \textit{transferability} of our proposed \method on a range of generative and discriminative tasks, demonstrating its generalization ability.

\definecolor{myblue}{rgb}{0.6, 0.77, 1.0}
\begin{figure*}
    \includegraphics[width=1.0\linewidth]{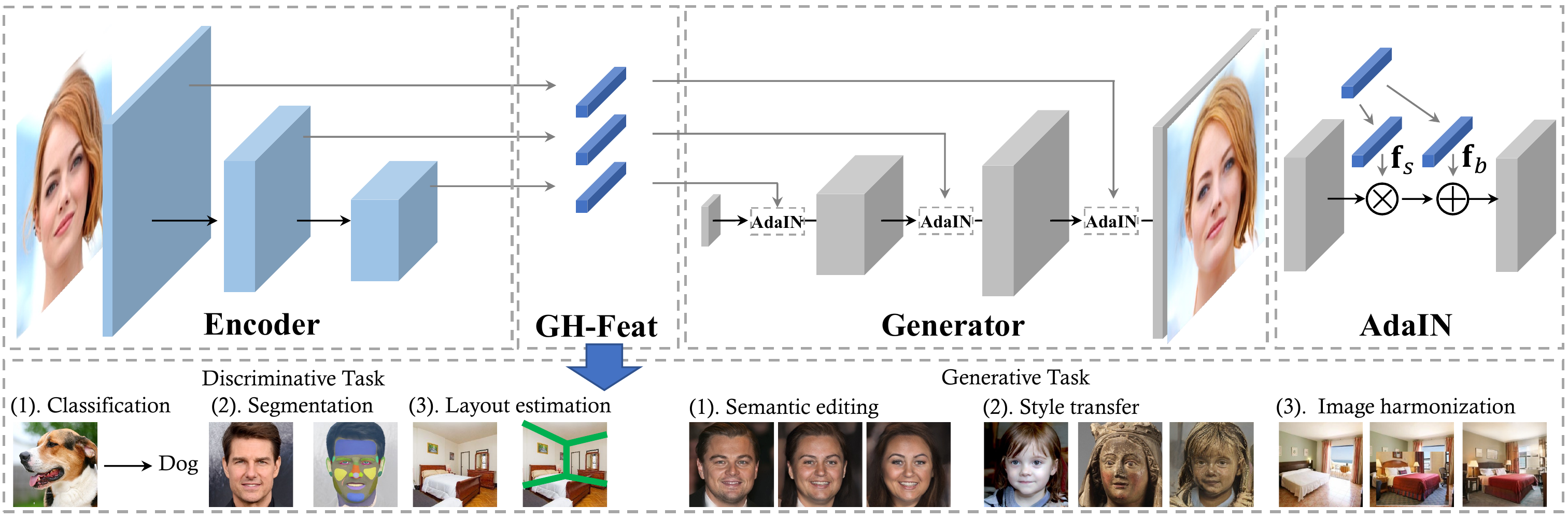}
    \vspace{-22pt}
      \caption{
        \textbf{Framework} of the GH-Feat.
        This feature hierarchy highly aligns with the layer-wise representation (\textit{i.e.}, style codes of per-layer AdaIN) learned by the StyleGAN generator.
        Parameters in \textbf{\textcolor{myblue}{blue}} blocks are learnable while others are frozen. 
      }
      \vspace{-5pt}
  \label{fig:framework}
\end{figure*}

\section{Methodology}\label{sec:method}
This section introduces the encoder used to extract hierarchical visual features from the input images.
This encoder is trained in an unsupervised manner using a well-prepared StyleGAN generator.
\cref{subsec:layerwise-representation} describes how we abstract the multi-level representation from StyleGAN.
\cref{subsec:hierarchical-encoder} presents the structure of the novel hierarchical encoder.
\cref{subsec:stylegan-as-loss} describes the idea of using pre-trained StyleGAN generator as a learned loss function for representation learning.
\cref{subsec:regularizer} introduces the training regularizer to prevent the encoder from overfitting pixel values.

\begin{table*}[t]
\begin{center}
  \caption{
    \textbf{Encoder structure}, which is based on ResNet-50~\cite{resnet}.
    Fully-connected (FC) layers are employed to map the feature maps produced by the Spatial Alignment Module (SAM) to our proposed Generative Hierarchical Features (GH-Feat).
    GH-Feat exactly align with the multi-scale style codes used in StyleGAN~\cite{stylegan}.
    The numbers in brackets indicate the dimension of features at each level.
  }
  \label{tab:arch}
  \vspace{-10pt}
  \small
  \begin{tabular}{ccccccc}
    \toprule
    Stage & Encoder Pathway &   Output Size & SAM \& Pool & FC Dimension & GH-Feat & Style Code in StyleGAN \\ 
    \midrule
    \multirow{2}{*}{input} & \multirow{2}{*}{$-$} &  \multirow{2}{*}{$3\times 256^2$} & & & & \\
    & & & & & & \\
    \midrule
    \multirow{2}{*}{conv$_1$} & \multicolumn{1}{c}{7$\times$7, {64}} & \multirow{2}{*}{$64\times 128^2$} & & & & \\
    & stride 2, 2 & & & & &\\
    \midrule
    \multirow{2}{*}{pool$_1$} & \multicolumn{1}{c}{3$\times$3, max} &  \multirow{2}{*}{$64\times 64^2$} & & & & \\
    & stride 2, 2 & & & & &\\
    \midrule
    \multirow{3}{*}{res$_2$} & \blocks{{256}}{{64}}{3} & \multirow{3}{*}{$256\times 64^2$} & & & & \\
    & & & & & & \\
    & & & & & & \\
    \midrule
    \multirow{3}{*}{res$_3$} & \blocks{{512}}{{128}}{4}  &  \multirow{3}{*}{$512\times 32^2$} & \multirow{3}{*}{$512\times 4^2$} &\multirow{3}{*}{8192$\times$1792} & Level 1-2 & Layer 14-13 ($128d\times2$) \\
    & & & & & Level 3-4  & Layer 12-11 ($256d\times2$) \\
    & & & & & Level 5-6  & Layer 10-9 ($512d\times2$) \\
    \midrule
    \multirow{3}{*}{res$_4$} & \blocks{{1024}}{{256}}{6} &  \multirow{3}{*}{$1024\times 16^2$} & \multirow{3}{*}{$512\times 4^2$} &\multirow{3}{*}{8192$\times$4096} & Level 7-8  & Layer 8-7 ($1024d\times2$) \\
    & & & & & Level 9-10  & Layer 6-5 ($1024d\times2$) \\
    & & & & & & \\
    \midrule
    \multirow{3}{*}{res$_5$} & \blocks{{2048}}{{512}}{3} & \multirow{3}{*}{$2048\times 8^2$} &  \multirow{3}{*}{$512\times 4^2$} &\multirow{3}{*}{8192$\times$4096} & Level 11-12  & Layer 4-3 ($1024d\times2$) \\
    & & & & & Level 13-14  & Layer 2-1 ($1024d\times2$) \\
    & & & & & & \\
    \bottomrule
  \end{tabular}
  \vspace{-15pt}
\end{center}
\end{table*}

\subsection{Layer-wise Representation from StyleGAN}\label{subsec:layerwise-representation}
\begin{figure}
    \includegraphics[width=1.0\linewidth]{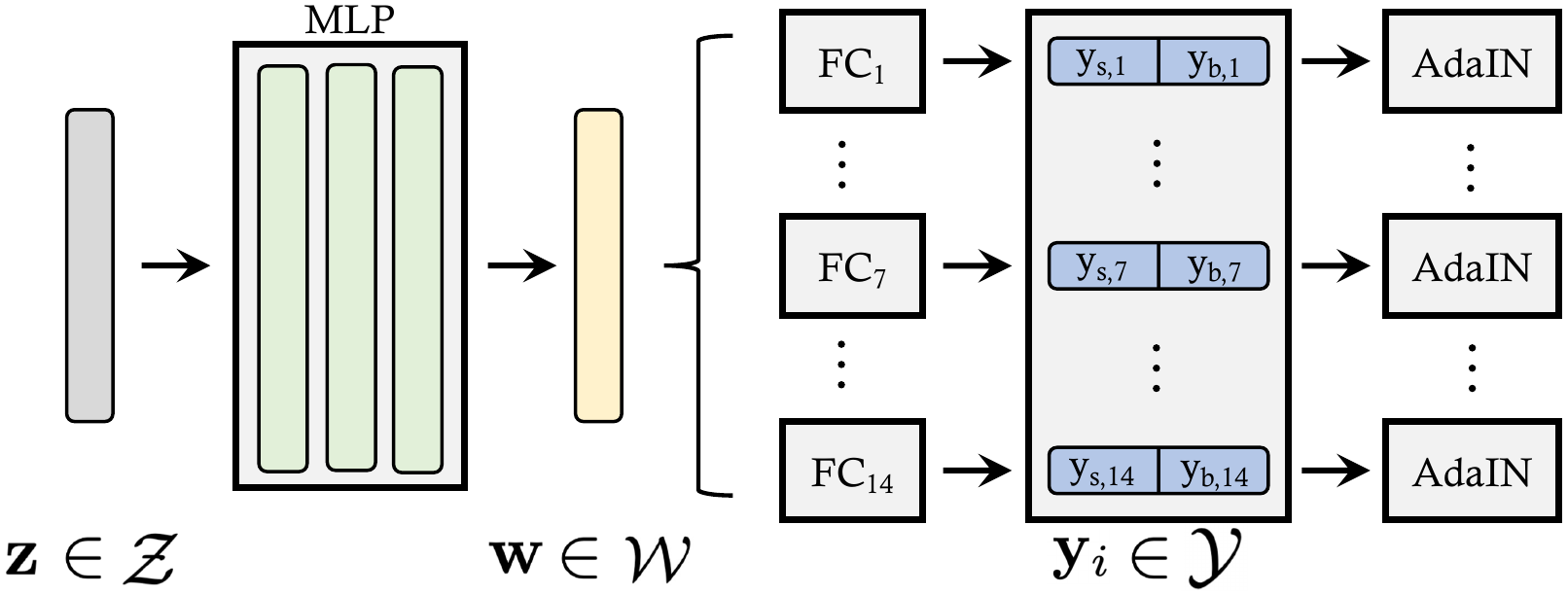}
    \vspace{-22pt}
      \caption{
        Multiple latent spaces of StyleGAN.
        %
        %
        FC refers to the affine layer between $\W$ space and $\Y$ space. 
      }
      \vspace{-5pt}
  \label{fig:latent-space}
\end{figure}

The generator $G(\cdot)$ of GANs typically takes a latent code $\z\in\Z$ as the input and is trained to synthesize a photo-realistic image $\x=G(\z)$.
The recent state-of-the-art StyleGAN~\cite{stylegan} proposes to first map $\z$ to a disentangled space $\W$ with $\w=f(\z)$.
Here, $f(\cdot)$ denotes the mapping implemented by multi-layer perceptron (MLP).
The $\w$ code is then projected to layer-wise style codes $\{\y^{(\ell)}\}_{\ell=1}^L \triangleq \{(\y_s^{(\ell)}, \y_b^{(\ell)})\}_{\ell=1}^L$ with affine transformations, where $L$ is the number of convolutional layers.
$\y_s^{(\ell)}$ and $\y_b^{(\ell)}$ correspond to the \revise{channel-wise} scale and weight parameters in Adaptive Instance Normalization (AdaIN)~\cite{adain}.
\revise{The space constructed by these layer-wise style parameters is named as $\Y$ space.}
These style codes are used to modulate the output feature maps of each convolutional layer with
\begin{align}
  \mathtt{AdaIN}(\x_i^{(\ell)}, \y^{(\ell)}) = \y_{s,i}^{(\ell)}\ \frac{\x_i^{(\ell)} - \mu(\x_i^{(\ell)})}{\sigma(\x_i^{(\ell)})} + \y_{b,i}^{(\ell)}, \label{eq:adain}
\end{align}
where $\x_i^{(\ell)}$ indicates the $i$-th channel of the output feature map from the $\ell$-th layer.
$\mu(\cdot)$ and $\sigma(\cdot)$ denote the mean and variance respectively.
\revise{\cref{fig:latent-space} illustrates the $\Z$, $\W$ and $\Y$ space of StyleGAN.}

Here, we treat the layer-wise style codes \revise{of $\Y$ space}, $\{\y^{(\ell)}\}_{\ell=1}^L$, as the generative visual features that we would like to extract from the input image.
There are two major advantages.
First, the synthesized image can be completely determined by these style codes without any other variations, making them suitable to express the information contained in the input data from the generative perspective.
Second, these style codes are organized as a hierarchy where codes at different layers correspond to semantics at different levels~\cite{stylegan,higan}.
To the best of our knowledge, this is the first work that adopts the style codes for the per-layer AdaIN module as the learned representations of StyleGAN.
\revise{Wu~\textit{et al.}~\cite{wu2021stylespace} also shows $\Y$ space can be leveraged for disentangled control of image editing, while our work explores the potential of generative representations in facilitating both generative and more importantly discriminative downstream tasks.}

\subsection{Hierarchical Encoder}\label{subsec:hierarchical-encoder}
Based on the layer-wise representation described in \cref{subsec:layerwise-representation}, we propose a novel encoder $E(\cdot)$ with a hierarchical structure to extract multi-level visual features from a given image. 
As shown in \cref{fig:framework}, the encoder is designed to best align with the StyleGAN generator.
In particular, the Generative Hierarchical Features (\method) produced by the encoder, $\{\f^{(\ell)}\}_{\ell=1}^L \triangleq \{(\f_s^{(\ell)}, \f_b^{(\ell)})\}_{\ell=1}^L$, are fed into the per-layer AdaIN module of the generator by replacing the style code $\y^{(L-\ell+1)}$ in Eq.~\eqref{eq:adain}.

We adopt ResNet~\cite{resnet} architecture as the encoder backbone and add an extra residual block to get an additional feature map with lower resolution.%
In fact, there are totally six stages in our encoder, where the first one is a convolutional layer (followed by a pooling layer) and each of the others consists of several residual blocks.
Besides, we introduce a feature pyramid network~\cite{fpn} to learn the features from multiple levels.
The output feature maps from the last three stages, $\{R_4, R_5, R_6\}$, are used to produce \method.
Taking a 14-layer StyleGAN generator as an instance, $R_4$ aligns with layer 9-14, $R_5$ with 5-8, while $R_6$ with 1-4.
Here, to bridge the feature map with each style code, we first downsample it to $4\times4$ resolution and then map it to a vector of the target dimension using a fully-connect (FC) layer.
In addition, we introduce a lightweight Spatial Alignment Module (SAM)~\cite{tpn, parnet} into the encoder structure to better capture the spatial information from the input image.
SAM works in a simple yet efficient way:
\begin{align}
  R_i &= W_i\mathtt{down}(R_i) + W_6R_6 \quad i \in \{4,5\}, \nonumber
\end{align}
where $W_4$, $W_5$, and $W_6$ (all are implemented with an $1\times1$ convolutional layer) are used to project the feature maps $R_4$, $R_5$, and $R_6$ to have the same number of feature channels respectively.
$R_4$ and $R_5$ are downsampled to the same resolution of $R_6$ before fusion.
%

\noindent{\textbf{Encoder Structure}}.
\cref{tab:arch} provides the detailed architecture of our hierarchical encoder by taking a 14-layer StyleGAN~\cite{stylegan} generator as an instance.
Recall that the design of \method treats the layer-wise style codes used in the StyleGAN model (\textit{i.e.}, the code fed into the AdaIN module~\cite{adain}) as generative features.
Accordingly, \method consists of 14 levels that exactly align with the multi-scale style codes yet in a reverse order, as shown in the last two columns of \cref{tab:arch}.

\subsection{Statistical Training Regularizer}\label{subsec:regularizer}

As discussed in \cref{subsec:layerwise-representation}, our approach aims at learning the style representations encoded in $\y$, which are transformed from the $\w$ code using pre-layer linear projection.
$\Y$ space is less constrained than $\W$ space and hence may suffer from the problem of overfitting pixel values, which further leads to poor transferability of the learned features.
To solve such a problem, we infer $\{\y_{avg}^{(\ell)}\}_{\ell=1}^L$ from the averaged latent code (\textit{i.e.}, a statistics from the training stage), $\w_{avg}$, and propose to only learn the residual code at each layer.
Thus, we have $E(\x) = \{\Delta\y^{(\ell)}\}_{\ell=1}^L$, which induces the final features as $\{\y_{avg}^{(\ell)} + \Delta\y^{(\ell)}\}_{\ell=1}^L$.
We then penalize the $l_2$ norm of each residual code to prevent it from shifting too far from the native distribution, resulting in a training regularizer
\begin{align}
    \mathcal{L}_{reg} = \sum_{\ell=1}^{L}\|\Delta\y^{(\ell)}\|_2^2.
\end{align}

\revise{e4e~\cite{tov2021e4e} also regularizes the inversion space when training the encoder yet from a different aspect against \method.
In particular, the regularization in e4e~\cite{tov2021e4e} targets minimizing the latent code variation across layers (\textit{i.e.}, they expect the inverted codes regarding different layers to be close to each other) to reconstruct the input image from coarse to fine.
Differently, the regularization in our work bonds the latent code close to the distribution center (\textit{i.e.}, the statistical average) to prevent the model from overfitting pixel values.
In this way, our approach could better represent an image from the semantic level, further facilitating downstream tasks.}

\subsection{StyleGAN Generator as Learned Loss}\label{subsec:stylegan-as-loss}
We consider the pre-trained StyleGAN generator as a leaned loss function.
Specifically, we employ a StyleGAN generator to supervise the encoder training with the objective of image reconstruction.
We also introduce a discriminator to compete with the encoder, following the formulation of GANs~\cite{gan}, to ensure the reconstruction quality.
To summarize, the encoder $E(\cdot)$ and the discriminator $D(\cdot)$ are jointly trained with
\begin{align}
  \begin{split}
    \min_{\Theta_E}\mathcal{L}_E =\ &||\x - G(E(\x))||_2 - \lambda_1 \E_{\x}[D(G(E(\x)))] \\
                                    &+ \lambda_2 ||F(\x) - F(G(E(\x)))||_2 + \lambda_3 \mathcal{L}_{reg}, \label{eq:encoder} 
  \end{split} \\
  \begin{split}
    \min_{\Theta_D}\mathcal{L}_D =\ &\E_{\x}[D(G(E(\x)))] - \E_{\x}[D(\x)] \\
                                    &+ \lambda_4 \E_{\x}[||\nabla_{\x}D(\x)||_2^2], \label{eq:discriminator}
  \end{split}
\end{align}
where $||\cdot||_2$ denotes the $l_2$ norm and $\lambda_1, \lambda_2, \lambda_3, \lambda_4$ are loss weights to balance different loss terms.
The last term in Eq.~\eqref{eq:encoder} represents the perceptual loss~\cite{johnson2016perceptual} and $F(\cdot)$ denotes the $\mathtt{conv4\_3}$ output from a pre-trained VGG~\cite{vgg} model.

\section{Experiments}\label{sec:experiments}
We evaluate Generative Hierarchical Features (\method) on a wide range of downstream applications.
\cref{exp:setting} introduces the experimental settings, such as implementation details, datasets, and tasks.
\cref{exp:analysis} presents the analysis of our approach including ablation study and the importance of the regularizer.
%
%
\cref{exp:generative} and \cref{exp:discriminative} evaluate the applicability of \method on generative and discriminative tasks respectively.
\cref{exp:spatial-task} shows the results of the introduced spatial expansion.

\subsection{Experimental Settings}\label{exp:setting}

\noindent\textbf{Implementation Details.}
The loss weights are set as $\lambda_1=0.1$, $\lambda_2=5e^{-5}$, $\lambda_3=5e^{-4}$ and $\lambda_4=5$.
We use Adam~\cite{adam} optimizer, with $\beta_1=0$ and $\beta_2=0.99$, to train both the encoder and the discriminator.
The learning rate is initially set as $1e^{-4}$ and exponentially decayed with the factor of 0.8.

\noindent\textbf{Datasets and Models.}
We conduct experiments on four StyleGAN~\cite{stylegan} models, pre-trained on MNIST~\cite{mnist}, FF-HQ~\cite{stylegan}, LSUN bedrooms~\cite{lsun}, and ImageNet~\cite{imagenet} respectively.
The MNIST model is with $32\times32$ resolution and the remaining models are with $256\times256$ resolution.

%
%

\noindent\textbf{Generative Tasks.}
\textit{\textbf{(1)} Image editing.}
It focuses on manipulating the image content or style, \textit{e.g.}, global editing, local editing.
\textit{\textbf{(2)} Image harmonization.}
This task harmonizes a discontinuous image to produce a realistic output.
\textit{\textbf{(3)} Style transfer.} This task focuses on transferring the style of the reference image to the source image.
\textit{\textbf{(4)} Semantic manipulation.} It targets at modifying the semantic meaning of an object while preserving other characteristics.
\textit{\textbf{(5)} Image colorization.} It focuses on colorizing the grayscale image.
\textit{\textbf{(6)} Image inpainting.} This task reconstructs missing regions in an image.
\textit{\textbf{(7)} Image super-resolution.} It aims at improving the resolution of the image.

\noindent\textbf{Discriminative Tasks.}
\textit{\textbf{(1)} MNIST digit recognition.}
It is a long-standing image classification task.
We report the Top-1 accuracy on the test set following~\cite{mnist}.
\textit{\textbf{(2)} Face verification.}
It aims at distinguishing whether the given pair of faces come from the same identity.
We validate on the LFW dataset~\cite{lfw} following the standard protocol~\cite{lfw}.
\textit{\textbf{(3)} ImageNet classification.}
This is a large-scale image classification dataset~\cite{imagenet}, consisting of over $1M$ training samples across 1,000 classes and $50K$ validation samples.
We use Top-1 accuracy as the evaluation metric following existing work~\cite{bigan,bigbigan}.
\textit{\textbf{(4)} Pose estimation.}
This task targets at estimating the yaw pose of the input face.
$70K$ real faces on FF-HQ~\cite{stylegan} are split into $60K$ training samples and $10K$ test samples.
The $\ell_1$ regression error is used as the evaluation metric.
\textit{\textbf{(5)} Landmark detection.}
This task learns a set of semantic points with visual meaning.
We use FF-HQ~\cite{stylegan} dataset and follow the standard MSE metric~\cite{tcdcn} to report performances in inter-ocular distance (IOD).
\textit{\textbf{(6)} Layout prediction.}
We extract the corner points of the layout line and convert the task to a landmark regression task.
The annotations of the collected $90K$ bedroom images ($70K$ for training and $20K$ for validation) are obtained with~\cite{layoutlearinng}.
Following~\cite{zou2018layoutnet}, we report the corner distance as the metric.
\textit{\textbf{(7)} Face luminance regression.}
It focuses on regressing the luminance of face images.
We use it as a low-level task on the FF-HQ~\cite{stylegan} dataset.
\textit{\textbf{(8)} Image retrieval.}
It aims at retrieving the images with specific attributes.
\textit{\textbf{(9)} Data-efficient image segmentation.}
This task focuses on predicting the class of each spatial pixel with limited annotated data.

\setlength{\tabcolsep}{6pt}
\begin{table}[t]
  \caption{
    Ablation studies on the feature space and the SAM module.
  }
  \vspace{-10pt}
  \label{tab:ablation}
  \centering\small
  \begin{tabular}{ccccccc}
    \toprule
    Space &       SAM & \revise{Reg} &  \MSE & \SSIM &  \FID \\ \midrule
    $\W$  & \ding{51} &     &    0.0601 & 0.540 & 22.24 \\
    $\Y$  &           &     & 0.0502 & 0.550 & 19.06 \\
    $\Y$  & \ding{51} &     & 0.0464 & 0.558 & 18.48 \\ 
    $\Y$  & \ding{51} & \revise{\ding{51}}  & \revise{0.0494} & \revise{0.551}  & \revise{16.84} \\
    
    \bottomrule
  \end{tabular}
\vspace{-10pt}
\end{table}

\subsection{Analysis on \method}\label{exp:analysis}

\subsubsection{Ablation Study}\label{exp:ablation}

We make ablation studies on the training of encoder from two perspectives.
(1) We choose the layer-wise style codes $\y$ over the $\w$ codes as the representation from StyleGAN.
(2) We introduce Spatial Alignment Module (SAM) into the encoder to better handle the spatial information.
(3) We involve a regularizer in the training of encoder.

Since the encoder is trained with the objective of image reconstruction, we use mean square error (MSE), SSIM~\cite{ssim}, and FID~\cite{fid} to evaluate the encoder performance.
\cref{tab:ablation} shows the results where we can tell that our encoder benefits from the effective SAM module and that choosing an adequate representation space (\textit{i.e.}, the comparison between the first row and the last row) results in a better reconstruction.
\revise{Introducing the regularizer alleviates the pixel value overfitting and improves the reconstruction quality at the distribution level.}
More discussion on the differences between $\W$ space and $\Y$ space can be found in \cref{exp:hierarchical}.

\begin{figure}[t]
  \centering
  \includegraphics[width=1.0\linewidth]{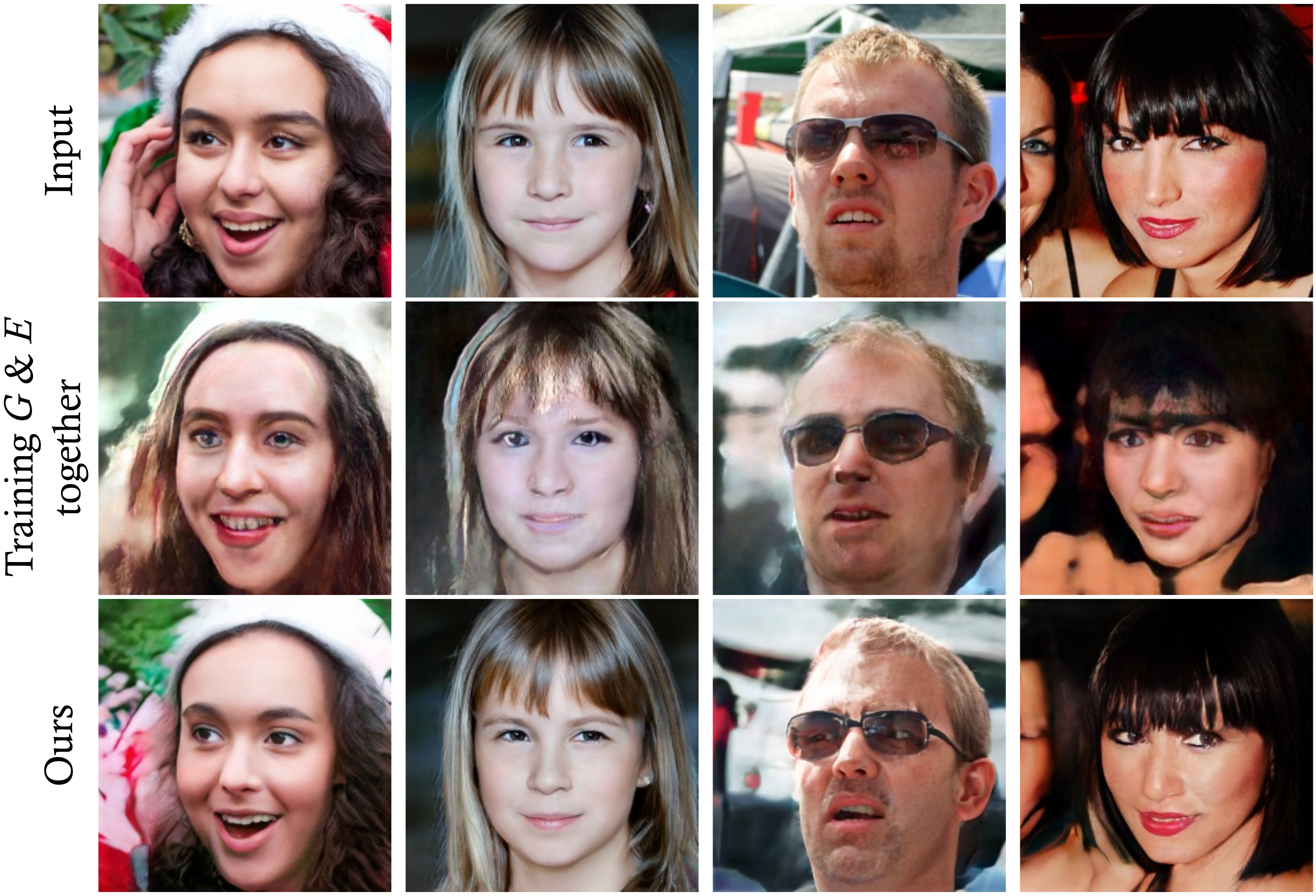}
    \vspace{-22pt}
  \caption{
    Qualitative comparison on image reconstruction between training the generator from scratch together with the encoder, and our \method that treats the well-learned StyleGAN generator as a loss function.
    }
  \label{supp:fig:random_g}
  \vspace{-8pt}
\end{figure}

\noindent{\textbf{Random Generator.}}
Recall that, during the training of the encoder, we propose to treat the well-trained StyleGAN generator as a learned loss function.
In this part, we explore what will happen if we train the generator from scratch together with the encoder.
\cref{supp:fig:random_g} and \cref{supp:tab:random_g} show the qualitative and quantitative results respectively, which demonstrate the strong performance of \method.
It suggests that besides higher efficiency, reusing the knowledge from a well-trained generator can also bring better performance.

\setlength{\tabcolsep}{8pt}
\begin{table}[t]
  \caption{
    Quantitative comparison on image reconstruction between training the generator from scratch together with the encoder, and our \method that treats the well-learned StyleGAN generator as a loss function. \revise{\method-R denotes \method trained with regularizer.}
  }
  \vspace{-10pt}
  \label{supp:tab:random_g}
  \centering\small
  \begin{tabular}{lccc}
    \toprule
                                     &   \MSE & \SSIM &  \FID \\ \midrule
    Training $G(\cdot)$ from Scratch &  0.429 & 0.301 & 46.20 \\
    \method (Ours)                   & 0.0464 & 0.558 & 18.48 \\ 
    \revise{\method-R}                 & \revise{0.0494} & \revise{0.551}  & \revise{16.84} \\
    \bottomrule
  \end{tabular}
  \vspace{-10pt}
\end{table}

\setlength{\tabcolsep}{15pt}
\begin{table}[t]
    \centering
    \caption{Cosine similarity of the encoder output and the native latent code.}
    \vspace{-10pt}
    \begin{tabular}{ccc}
    \toprule
            &  \textit{w/o} $\mathcal{L}_{reg}$  & \textit{w/} $\mathcal{L}_{reg}$ \\ \midrule
    FED     & 0.444 & 0.879 \\
    \bottomrule
    \end{tabular}
    \label{tab:cos_sim}
    \vspace{-5pt}
\end{table}

\begin{figure*}[t]
  \centering
  \includegraphics[width=1.0\linewidth]{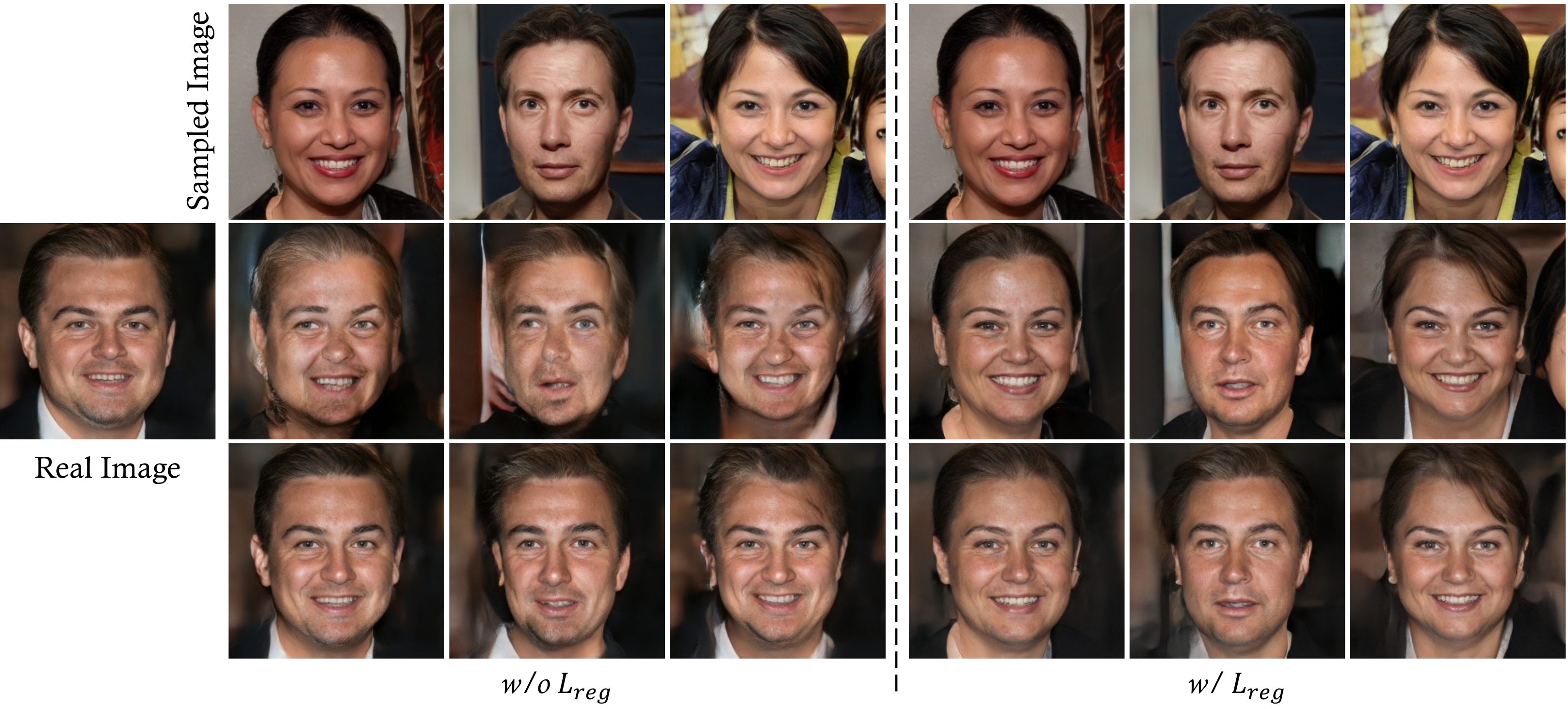}
  \vspace{-22pt}
  \caption{Qualitative comparison on the style mixing task between using training regularizer or not. The first row are the sampled images. The second row shows the results mixed with the codes predicted by our encoder.
  The third row presents the mixing results with original latent code.
    }
  \label{fig:stylemixing_comp}
  \vspace{-10pt}
\end{figure*}

\begin{figure*}[t]
  \centering

  \includegraphics[width=1.0\linewidth]{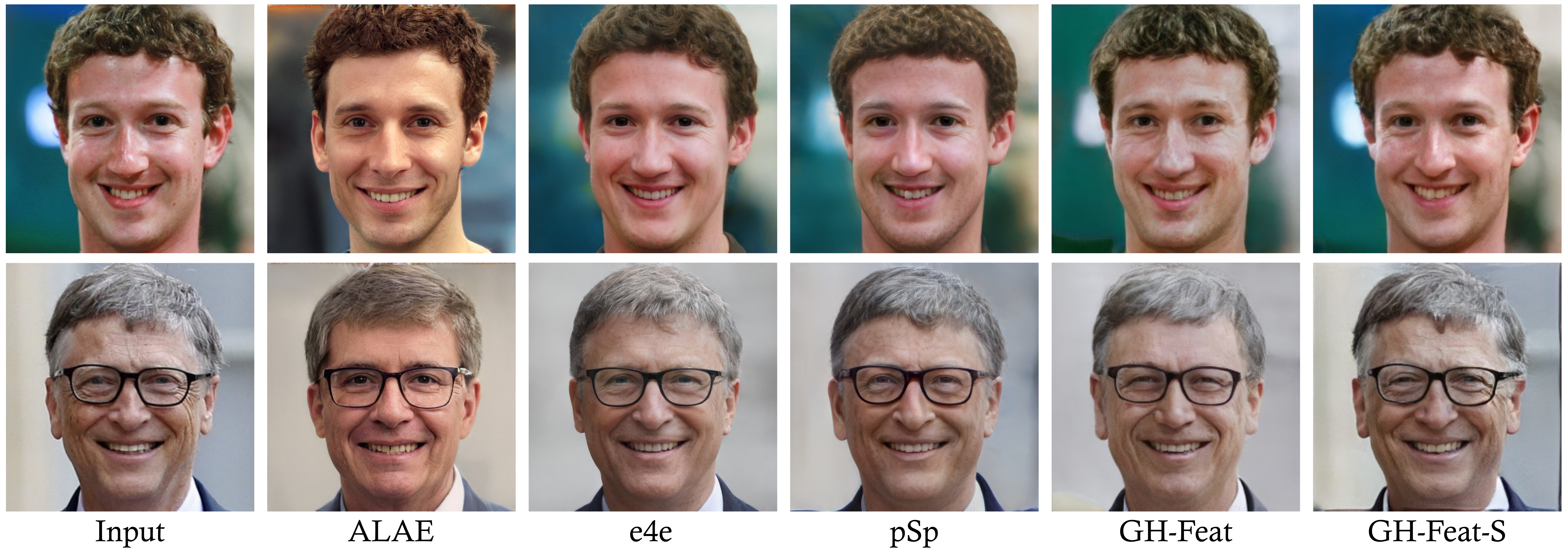}
  \vspace{-20pt}
  \caption{
    Qualitative comparison on reconstructing real images.
    \revise{\method-S denotes the spatial GH-Feat.
    Our \method and \method-S, which are built on StyleGAN, could get comparable and better performance as pSp~\cite{richardson2021pSp} and e4e~\cite{tov2021e4e}, which employ a more powerful StyleGAN2 generator.}
  }
  \label{fig:inversion}
  \vspace{-5pt}
\end{figure*}

\subsubsection{Importance of Regularizer} \label{exp:reg}
Although \method has achieved good results in image reconstruction, it cannot perform very well on image editing. 
Compared with the $\W$ space that previous attempts adopt as the inversion space, the $\Y$ space used by \method ignores the linear transformation between $\w$ code and $\y$ code, resulting in its flexibility.
Hence, it is easier to overfit a given image through a simple combination of generative features. 
This leads to a mismatch between the learned visual features and the latent space distribution of the generator.

We choose the task of global editing as a benchmark to explore the mismatch problem.
Specifically, we extract the generative features of the real image first and then randomly replace them with the randomly sampled features in the latent space at layers 0-4 to achieve global editing.
The results are shown in the 2nd row of \cref{fig:stylemixing_comp}.
Besides, we also extract the visual feature of the sampled images and do the same operation to achieve the editing result in the third line of \cref{fig:stylemixing_comp}. 
Obviously, the mixed results by the two sets of features both extracted by the encoder are better, suggesting the domain shift between the visual features and the latent space of the generator.
Based on this, we apply the constraints proposed in \cref{sec:method} to the encoder training. 
The right part of \cref{fig:stylemixing_comp} shows the editing results with the $\mathcal{L}_{reg}$. 
The global editing results with sampled and extracted features are very similar, and both are much better than the result without $\mathcal{L}_{reg}$.
It demonstrates that the generative features learned with $\mathcal{L}_{reg}$ are more in line with its distribution of latent space.

To quantitatively measure the similarity between two domains, we use cosine distance between generative feature and native code.
Specifically, we sample 10k fake images and extract the corresponding \method by our encoder, and then cosine similarity is calculated for the two distributions.  
As shown in \cref{tab:cos_sim}, minimizing the variation of the generative features can improve the similarity from 0.444 to 0.879, suggesting the effectiveness of this regularization.

\setlength{\tabcolsep}{13pt}
\begin{table*}[!ht]
  \caption{
    Quantitative comparison on reconstructing images from FF-HQ faces~\cite{stylegan} and LSUN bedrooms~\cite{lsun}. \revise{\method-S denotes the spatial \method. \textbf{bold} ones rank the best among the methods w/o generator tuning and \underline{underlined} ones are the second.}
  }
  \label{tab:inversion}
  \vspace{-10pt}
  \centering\small
  \begin{tabular}{lccccc}
    \toprule
            & \multicolumn{3}{c}{Face} & \multicolumn{2}{c}{Bedroom} \\ \cmidrule[0.5pt](lr){2-4} \cmidrule[0.5pt](lr){5-6}
    Method           &  \MSE & \SSIM & \TIME & \MSE & \SSIM  \\ \midrule
    \multicolumn{6}{c}{\textit{w/ generator tuning}}                    \\[2pt]
    PTI~\cite{roich2021pivotal} & 0.009 & 0.74  & 58.02  & -  & - \\ \midrule
    \multicolumn{6}{c}{\textit{w/o generator tuning}}                    \\[2pt]
    ALAE~\cite{alae} & 0.182 & 0.40 & \textbf{0.023} & 0.275 & 0.32  \\

    pSp~\cite{richardson2021pSp} & 0.034 & 0.56 & 0.063 & - & -  \\
    e4e~\cite{tov2021e4e}        & 0.052 & 0.50  &0.063 & - & -  \\
    Restyle~\cite{alaluf2021restyle} & \underline{0.030} & \underline{0.66} & 0.304 & - & -  \\
    \method   & 0.046 & 0.56 & 0.035 & 0.068 & 0.52  \\

    \method-S   & \textbf{0.029} & \textbf{0.67} & 0.038 & \textbf{0.057} & \textbf{0.581}\\
    \bottomrule 
  \end{tabular}
  \vspace{-5pt}
\end{table*}

\begin{figure*}[t]
  \centering
  \includegraphics[width=1.0\linewidth]{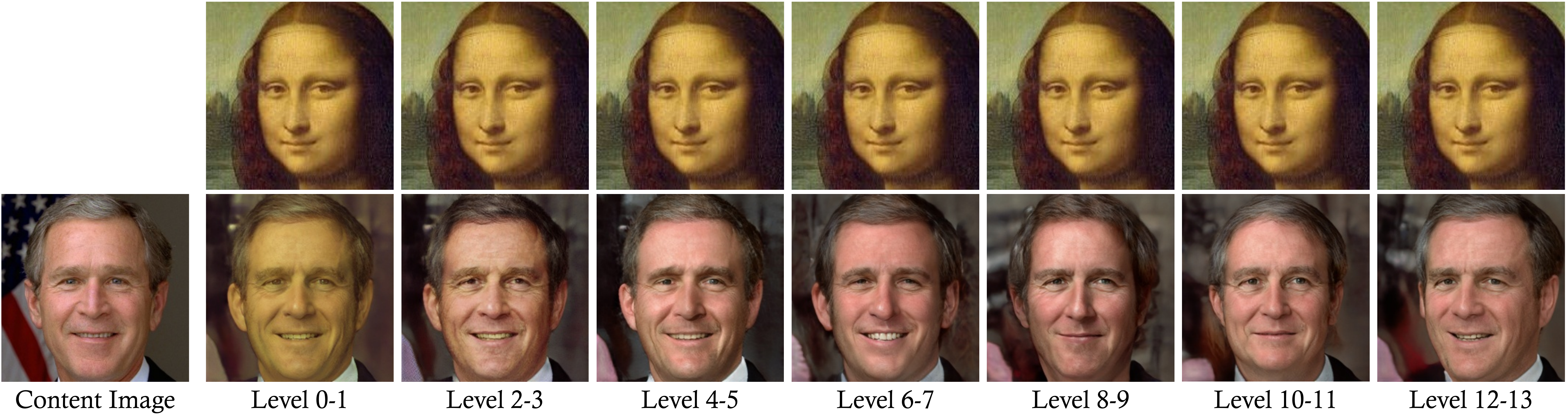}
    \vspace{-22pt}
  \caption{
    \textbf{Style mixing} results by exchanging the \method extracted from the content image and the style image (first row) at different levels.
    %
    \revise{Higher level corresponds to the high-level semantics like the face shape and pose information.}
  }
  \label{fig:style-mixing}
  \vspace{-7pt}
\end{figure*}

\begin{figure*}[t]
  \centering
  \includegraphics[width=1.0\linewidth]{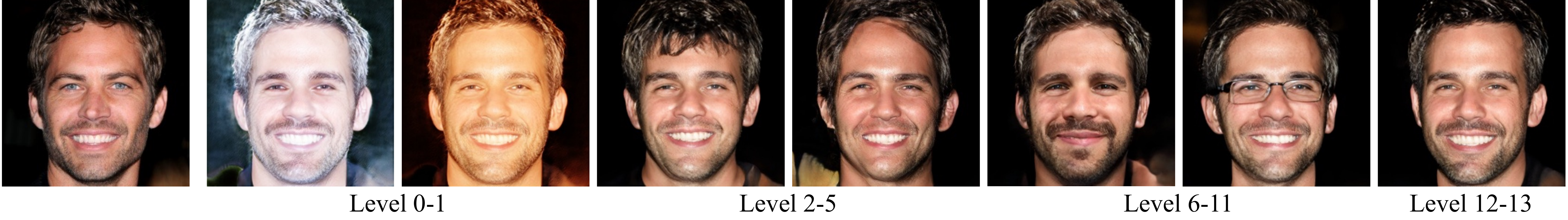}
    \vspace{-22pt}
  \caption{
    \textbf{Global image editing} achieved by \method.
    On the left is the input image, while the others are generated by randomly sampling the visual feature at some particular level.
  }
  \label{fig:global}
  \vspace{-7pt}
\end{figure*}

\begin{figure*}[t]
  \centering
  \includegraphics[width=1.0\linewidth]{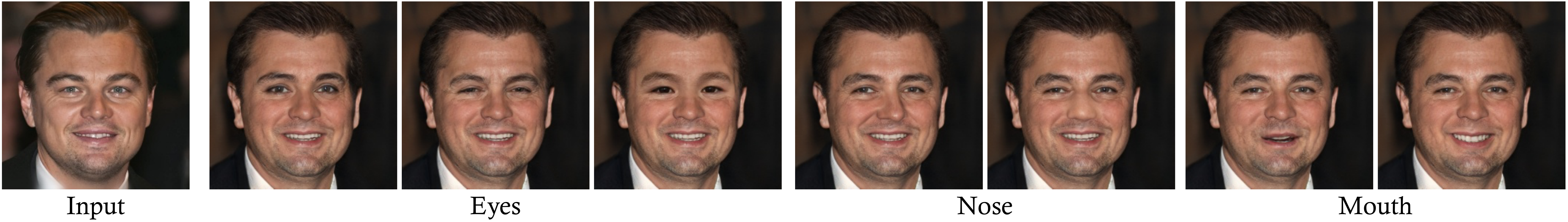}
    \vspace{-22pt}
  \caption{
    \textbf{Local image editing} achieved by \method.
    On the left is the input image, while the others are generated by randomly sampling the visual feature and replacing the spatial feature map (for different regions) at some particular level.
    Zoom in for details.
  }
  \label{fig:local}
  \vspace{-7pt}
\end{figure*}

\begin{figure*}[t]
  \centering
  \includegraphics[width=1.0\linewidth]{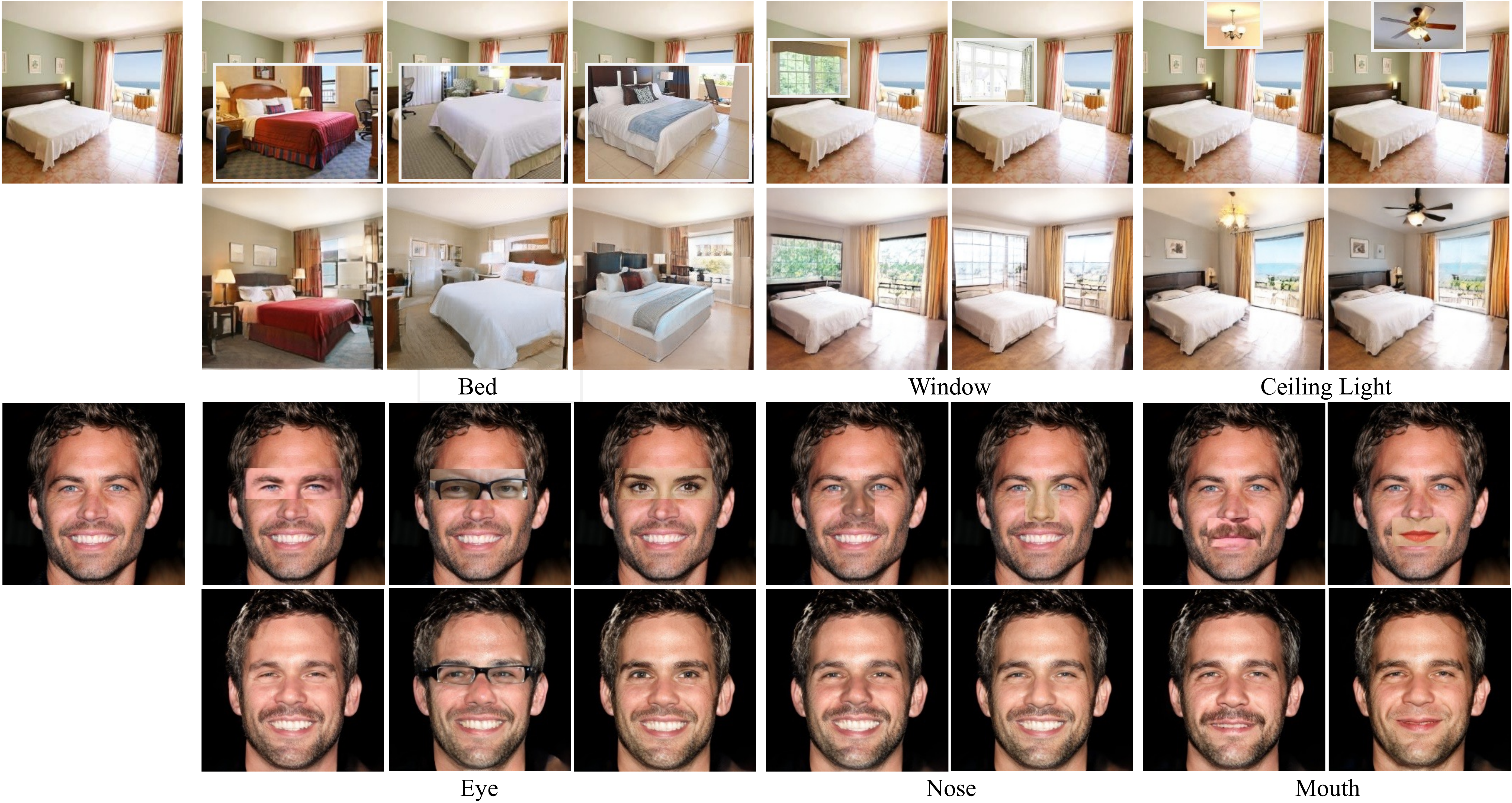}
    \vspace{-20pt}
  \caption{
    \textbf{Image harmonization} on bedroom and face with \method.
    \revise{The top left corner of the first and third rows} are the original images.
    Pasting a target image patch onto the original image then feeding it as the input (first and third row), our hierarchical encoder is able to smooth the image content and produce a photo-realistic image (second and fourth row).
  }
  \label{fig:harmonization}
  \vspace{-5pt}
\end{figure*}

\begin{figure*}[t]
  \centering
  \includegraphics[width=1.0\linewidth]{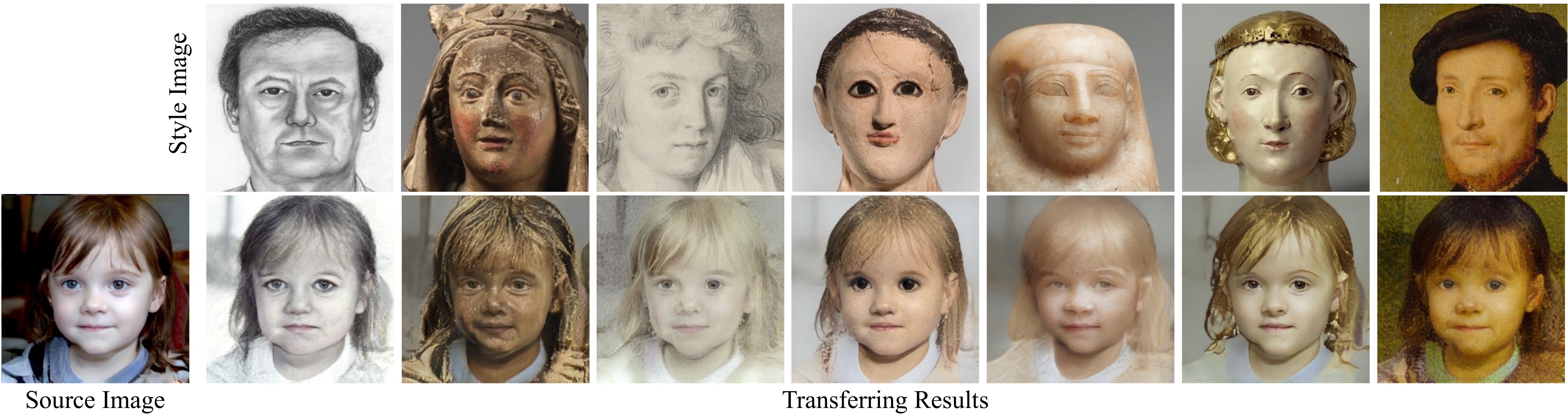}
    \vspace{-20pt}
  \caption{
    \textbf{Style transfer} results with \method. \method can extract and then transfer the style of the reference image to the given image.
    }
  \label{fig:conditional_synthesis}
  \vspace{-5pt}
\end{figure*}

\begin{figure*}[t]
  \centering
  \includegraphics[width=1.0\linewidth]{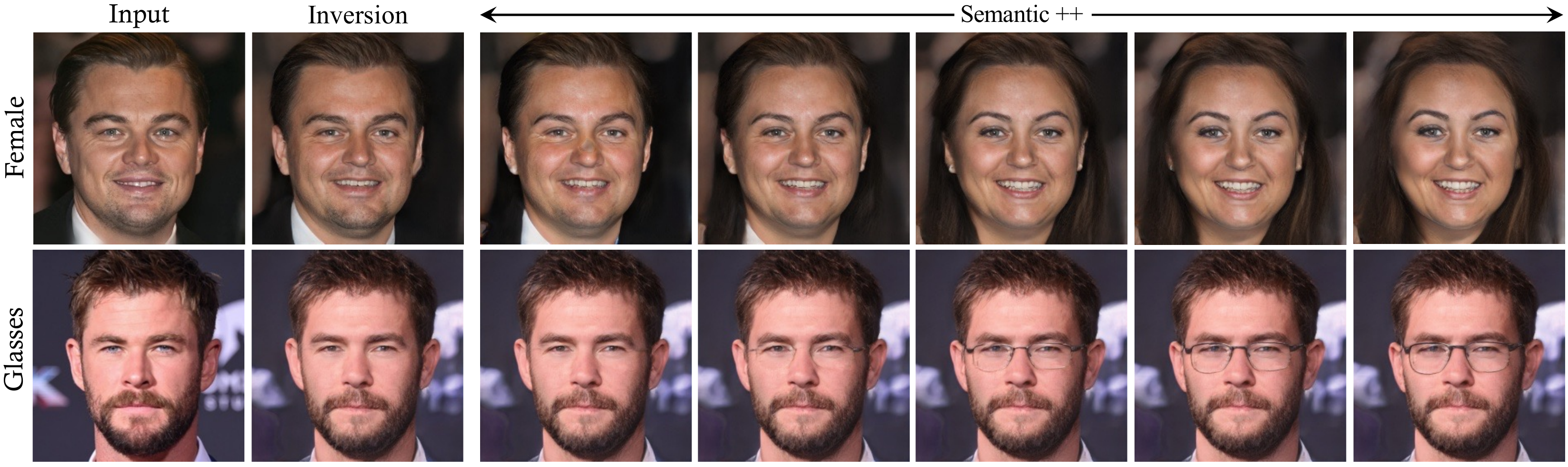}
    \vspace{-20pt}
  \caption{
    \textbf{Semantic Manipulation} results with \method. We utilize the off-the-shelf semantic directions from InterFaceGAN~\cite{interfacegan} to edit the gender and glasses of the given images.
    }
  \label{fig:semantci_editing}
  \vspace{-5pt}
\end{figure*}

\subsection{Evaluation on Generative Tasks}\label{exp:generative}

Thanks to using the StyleGAN as a learned loss function, a huge advantage of \method over existing unsupervised feature learning approaches~\cite{hjelm2019learning,zhuang2019local,cpc,cmc,moco}, which mainly focus on the image classification task, is its generative capability.
In this section, we conduct a number of generative experiments to verify this point.

\subsubsection{Image Reconstruction}\label{exp:reconstruction}
Image reconstruction is an important evaluation on whether the learned features can best represent the input image.
\revise{MSE and SSIM~\cite{ssim} are used as quantitative metrics to evaluate the reconstruction performance.}
\cref{tab:inversion} and \cref{fig:inversion} show the quantitative and qualitative comparison between our \method and other GAN inversion methods on FF-HQ faces~\cite{stylegan} and LSUN bedrooms~\cite{lsun}.
The very recent work ALAE~\cite{alae} also employs StyleGAN for representation learning.
We have following differences from ALAE:
(1) We use the $\Y$ space instead of the $\W$ space of StyleGAN as the representation space.
(2) We learn \textit{hierarchical} features that highly align with the per-layer style codes in StyleGAN.
(3) Our encoder can be \textit{efficiently} trained with a well-learned generator by treating StyleGAN as a loss function.
We can tell that \method better reconstructs the input by preserving more information, resulting a more expressiveness representation.

\revise{
 Besides pSp~\cite{richardson2021pSp}, e4e~\cite{tov2021e4e} and Restyle~\cite{alaluf2021restyle}, we include the results of PTI~\cite{roich2021pivotal} as well as the improved version of our \method (\textit{i.e.}, spatial expansion introduced in Sec.~4.5).
We also include the inference time to help evaluate the model efficiency.
We have three observations from the table below.
(1) Our \method, which is built on StyleGAN, could get comparable performance as pSp~\cite{richardson2021pSp} and e4e~\cite{tov2021e4e}, which employ a more powerful StyleGAN2 generator. We surmise that such an advantage originates from the replacement from $\W$ space to $\Y$ space.
(2) Restyle~\cite{alaluf2021restyle} (which requires iterative refinement) and PTI~\cite{roich2021pivotal} (which requires tuning of the weights of the generator) provide good reconstruction results but suffer from slow inference speed.
(3) Our improved version, \textit{i.e.}, Spatial \method, substantially improves the inversion quality without sacrificing the model efficiency, and achieves the best performance among all encoder-based methods without generator tuning.
}

\subsubsection{Image Editing}\label{exp:editing}
In this part, we evaluate \method on a number of image editing tasks.
Different from the features learned from discriminative tasks~\cite{resnet,moco}, our \method naturally supports sampling and enables creating new data.

\noindent\textbf{Style Mixing.}
To achieve style mixing, we use the encoder to extract visual features from both the content image and the style image and swap these two features at some particular level.
The swapped features are then visualized by the generator, as shown in \cref{fig:style-mixing}.
We can observe the compelling hierarchical property of the learned \method.
For example, by exchanging low-level features, only the image color tone and the skin color are changed.
Meanwhile, mid-level features controls the expression, age, or even hair styles.
Finally, high-level features correspond to the face shape and pose information (last two columns).

\noindent\textbf{Global Editing.}
The style mixing results have suggested the potential of \method in multi-level image stylization.
Sometime, however, we may not have a target style image to use as the reference.
Thanks to the design of the latent space in GANs~\cite{gan}, the generative representation naturally supports sampling, resulting in a strong creativity.
In other words, based on \method, we can arbitrarily sample meaningful visual features and use them for image editing.
\cref{fig:global} presents some high-fidelity editing results at multiple levels.
This benefits from the matching between the learned \method and the internal representation of StyleGAN.

\noindent\textbf{Local Editing.}
Besides global editing, our \method also facilitates editing the target image locally by deeply cooperating with the generator.
In particular, instead of directly swapping features, we can exchange a certain region of the spatial feature map at some certain level.
In this way, only a local patch in the output image will be modified while other parts remain untouched.
As shown in \cref{fig:local}, we can successfully manipulate the input face with different eyes, noses, and mouths.

\subsubsection{Image Harmonization}\label{exp:harmonize}
Our hierarchical encoder is robust such that it can extract reasonable visual features even from discontinuous image content.
%
\revise{We copy the patches from other images onto the original image} and feed the stitched image into our proposed encoder for feature extraction.
The extracted features are then visualized via the pre-trained generator, as in \cref{fig:harmonization}.
%
\revise{On the bedroom, we can see that the copied bed, window and ceiling light well blend into the ``background''.}
We also surprisingly find that when copying a window into the source image, the view from the original window and that from the new window highly align with each other (\textit{e.g.}, vegetation or ocean).
\revise{On face image, besides eye, nose and mouth, \method also blends the glasses with the background very well,  benefiting from the robust generative visual features.
}

\subsubsection{Style Transfer} \label{exp:styletransfer}

Our \method can not only edit the image attributes by replacing the randomly sampling feature at a particular level but also can facilitate the editing with the given conditional input. 
Here, we take style transfer as an example, aiming to transfer the style of the given image to the source image.
We first extract the generative features of the content image $I_c$ and style image $I_s$, and then style-mixing is performed by replacing the visual features of $I_c$ with the corresponding ones of $I_s$ at the layer 8-16.
We leverage the disentanglement of the generative features across different layers to perform style transfer.
As shown in \cref{fig:conditional_synthesis}, our encoder can successfully transfer the style of the given image to the source images, suggesting the effectiveness of the generative features. 
It is worth noting that although the texture of the given style images rarely appears in the training dataset,  our encoder can still reconstruct it and extract reasonable visual features with good disentangle properties.
It also supports the robustness and generalization of the visual features extracted by our hierarchical encoder.

\begin{figure*}[t]
  \centering
  \includegraphics[width=1.0\linewidth]{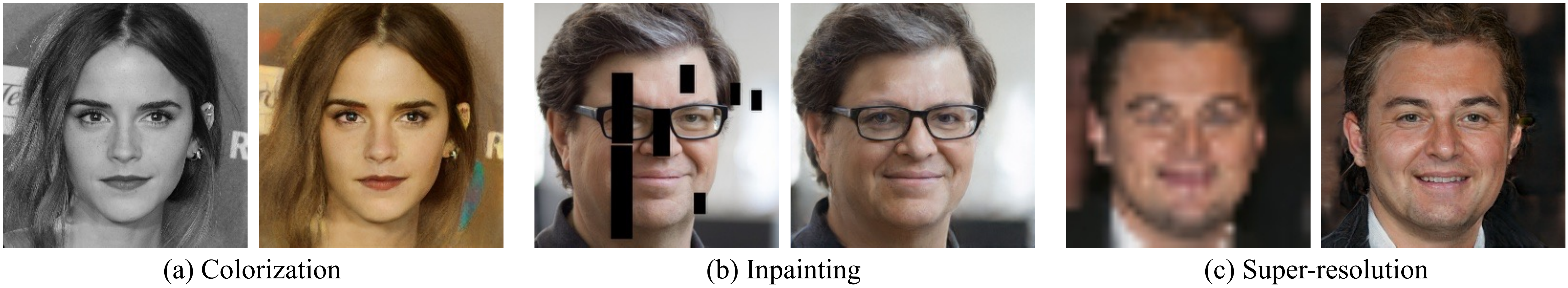}
    \vspace{-22pt}
  \caption{
    \textbf{Image processing} with \method.
    \method facilitates many image processing applications using the hierarchical encoder.
    }
  \label{fig:processing}
  \vspace{-7pt}
\end{figure*}

\begin{figure*}[t]
  \centering
  \includegraphics[width=1\linewidth]{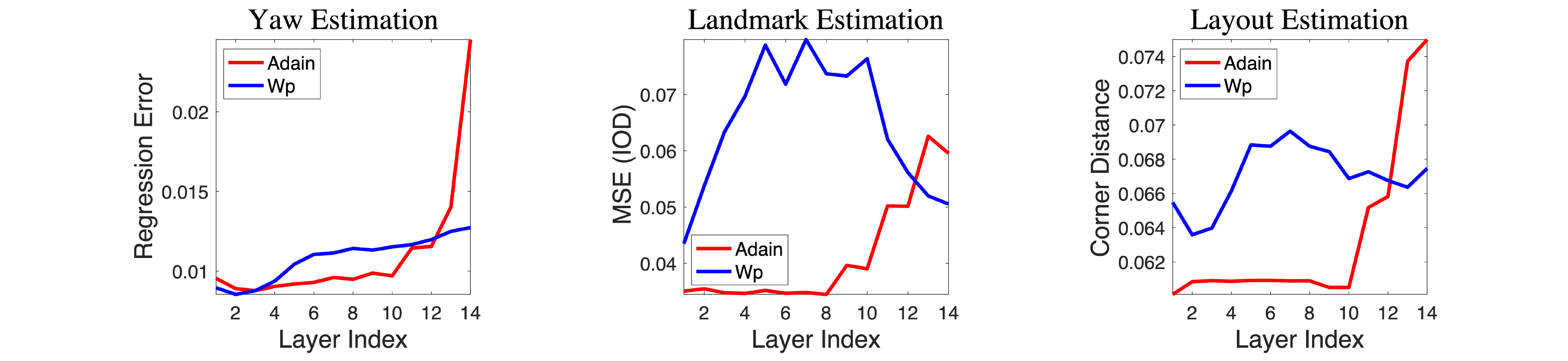}
     \vspace{-22pt}
  \caption{
    Performances on different discriminative tasks using \method.
    Left three columns enclose the comparisons between using different spaces of StyleGAN as the representation space, where $\Y$ space (in \textbf{\textcolor{red}{red}} color) shows stronger discriminative and hierarchical property than $\W$ space (in \textbf{\textcolor{blue}{blue}} color).
    This is discussed in \cref{exp:hierarchical}.
    The last column compares the two different strategies used in the face verification task, which is explained in \cref{exp:classification}.
    %
  }
  \label{fig:layerwise}
  \vspace{-7pt}
\end{figure*}

\subsubsection{Semantic Manipulation} \label{exp:semanticediting}
Here we explore the semantic editability of the generative features.
We utilize off-the-shelf semantic directions from InterFaceGAN
~\cite{interfacegan} to edit the inversion results.
\cref{fig:semantci_editing} presents the results of the manipulated faces.
Obviously, the learned generative features can preserve most other details when manipulating a particular facial attribute.
These editing results demonstrate that generative features can not only reconstruct the given image in high quality, but also facilitate it with good semantic manipulation properties.

\subsubsection{Image Processing} \label{exp:image-processing}
In this section, we demonstrate that our method facilitates various image processing tasks such as image colorization, image inpainting, and image super-resolution by utilizing the prior knowledge learned by GANs.
Generally, these tasks can be formulated as follows:

\begin{equation}
    s^{*} = \arg \min_{s \in S}L(G(s), x).
\end{equation}
where $ s $ is the style code initialized by our encoder, $ L $ is the l2 loss function, and $ x $ is the reference image (e.g., gray-scale image for image colorization, corrupted image for the inpainting, and low-resolution image for super-resolution).

Image colorization tries to restore the original color of a gray-scale image.
The results from our method are listed in \cref{fig:processing}a. 
Image inpainting aims at filling the missing pixels of the input images.
As shown in \cref{fig:processing}b, when some pixels value of the input image is missing, our method still successfully recovers them.
The last one is super-resolution, which manages to generate a high-resolution image of the low-resolution one.
\cref{fig:processing}c shows the super-resolution result scale 16 times using our method.

\subsection{Evaluation on Discriminative Tasks}\label{exp:discriminative}
In this part, we verify that even the proposed \method is learned from generative models, it can be applicable to a wide range of discriminative tasks with competitive performances.
Here, we do not fine-tune the encoder for any certain task.
In particular, we choose multi-level downstream applications, including image classification, face verification, pose estimation, layout prediction, landmark detection, and luminance regression.
For each task, we use our encoder to extract visual features from both the training and the test set.
A linear regression model (\textit{i.e.}, a fully-connected layer) is learned on the training set with ground-truth and then evaluated on the test set.
Besides, we include image retrieval as an addition discriminative task to verify the hierarchical property of \method, whose details are explained in \cref{exp:retrieval}.

\subsubsection{Discriminative and Hierarchical Property}\label{exp:hierarchical}
Recall that \method is a multi-scale representation learned by using StyleGAN as a loss function.
As a results, it consists of features from multiple levels, each of which correspond to a certain layer in the StyleGAN generator.
Here, we would to explore how this feature hierarchy is organized as well as how they can facilitate multi-level discriminative tasks, including face pose estimation, indoor scene layout prediction, and luminance\footnote{We convert images from RGB space to YUV space and use the mean value from Y space as the luminance.} regression from face images.
In particular, we evaluate \method on each task level by level.
As a comparison, we also train encoders by treating the $\w$ code, instead of the style code $\y$, as the representation.
From \cref{fig:layerwise}, we have three observations:
(1) \method is discriminative.
(2) Features at lower level are more suitable for low-level tasks (\textit{e.g.}, luminance regression) and those at higher level better aid high-level tasks (\textit{e.g.}, pose estimation).
(3) $\Y$ space demonstrates a more obvious hierarchical property than $\W$ space.
\revise{The comparison on hierarchical property between using regularizer or not is included at \textbf{Supplementay Material}.}

\subsubsection{Digit Recognition \& Face Verification}\label{exp:classification}
Image classification is widely used to evaluate the performance of learned representations~\cite{hjelm2019learning,zhuang2019local,moco,cpc,bigbigan}.
In this section, we first compare our proposed \method with other alternatives on a toy dataset, \textit{i.e.}, MNIST \cite{mnist}.
Then, we use a more challenging task, \textit{i.e.}, face verification, to evaluate the discriminative property of \method.

\noindent\textbf{MNIST Digit Recognition.}
We first show a toy example on MNIST following prior work~\cite{bigan,alae}.
We make a little modification to ResNet-18 like~\cite{pytorch-cifar10} which is widely used in literatures to handle samples from MNIST~\cite{mnist} in lower resolution.
The Top-1 accuracy is reported in \cref{tab:comparison}a.
Our \method outperforms ALAE~\cite{alae} and BiGAN~\cite{bigan} with $1.45\%$ and $1.92\%$, suggesting a stronger discriminative power.
Here, ResNet-18~\cite{resnet} is employed as the backbone structure for both MoCo~\cite{moco} and \method.

\begin{figure*}[t]
  \centering
  \includegraphics[width=1\linewidth]{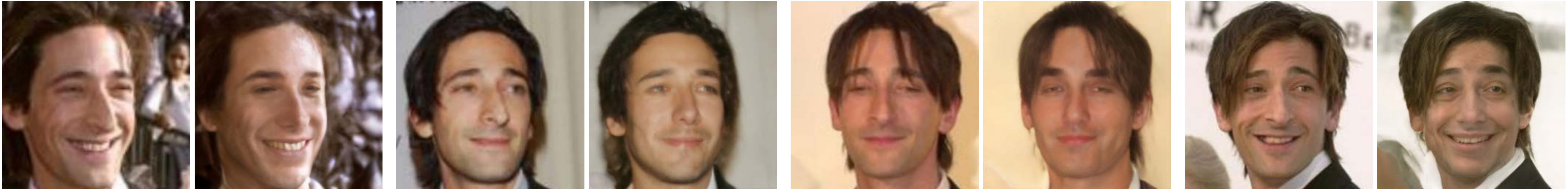}
  \vspace{-22pt}
  \caption{
    \textbf{Image reconstruction} results on LFW~\cite{lfw}.
    For each pair of images, left is the low-resolution input while right is reconstructed by \method.
    All samples are with the same identity.
  }
  \label{fig:lfw}
  \vspace{-7pt}
\end{figure*}

\begin{figure*}[t]
  \centering
  \includegraphics[width=1\linewidth]{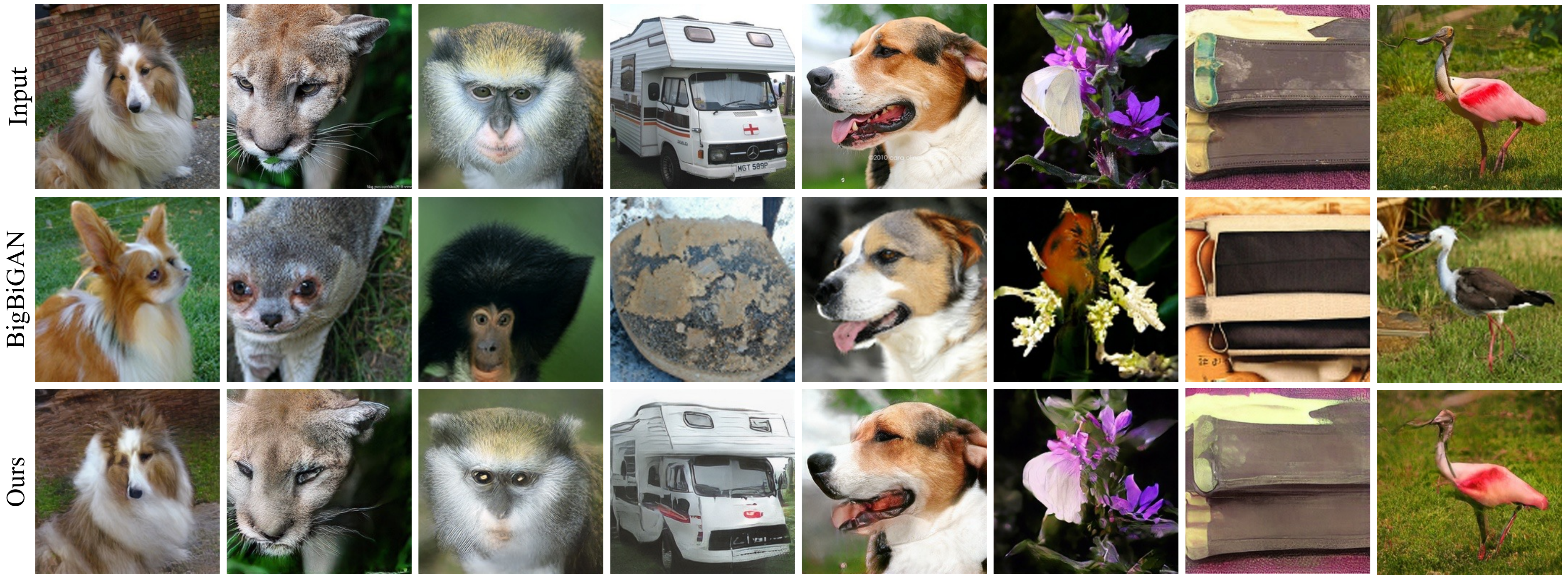}
  \vspace{-22pt}
  \caption{
    \textbf{Qualitative comparison} between BigBiGAN~\cite{bigbigan} and \method on reconstructing images from ImageNet~\cite{imagenet}.
  }
  \label{fig:imagenet}
  \vspace{-7pt}
\end{figure*}

\begin{table}[t]
  \captionsetup[subfloat]{captionskip=2pt,position=top}
  \caption{
    Quantitative comparison between our proposed \method and other alternatives on MNIST~\cite{mnist} and LFW~\cite{lfw}.
    \revise{\method-R denotes \method trained with regularizer.}
  }
  \label{tab:comparison}
   \vspace{-15pt}
  \centering
  \subfloat[Digit recognition on MNIST.]{
    \centering\small
    \setlength{\tabcolsep}{5.5pt}
    \begin{tabular}{lc}
      \toprule
      Methods                           &  Acc. \\ \midrule
      AE($\ell_{1}$)~\cite{autoencoder} & 97.43 \\
      AE($\ell_{2}$)~\cite{autoencoder} & 97.37 \\
      BiGAN~\cite{bigan}                & 97.14 \\
      ALAE~\cite{alae}                  & 97.61 \\
      MoCo-R18~\cite{moco}              & 95.89      \\ \midrule
      \method (Ours)                    & 99.06 \\ 
      \revise{\method-R}           &  \revise{98.78}     \\ 
      \bottomrule
    \end{tabular}
  }
  \hspace{2pt}
  \subfloat[Face verification on LFW.]{
    \centering\small
    \setlength{\tabcolsep}{4.5pt}
    \begin{tabular}{lc}
      \toprule
      Methods              & Acc. \\ \midrule
      VAE~\cite{vae}       & 49.3 \\
      MoCo-R50~\cite{moco} & 48.9 \\ 
      ALAE~\cite{alae}     & 55.7 \\ \midrule
      \method (Grouping)   & 60.1 \\
      \method (Layer-wise) & 67.5 \\
      \method (Voting)     & 69.7 \\ 
      \revise{\method-R (Voting)} & \revise{69.1}  \\ 
      \bottomrule
    \end{tabular}
  }
   \vspace{-10pt}
\end{table}

\noindent\textbf{LFW Face Verification.}
We directly use the proposed encoder, which is trained on FF-HQ~\cite{stylegan}, to extract \method from face images in LFW~\cite{lfw} and tries three different strategies on exploiting \method for face verification:
(1) using a single level feature;
(2) grouping multi-level features (starting from the highest level) together;
(3) voting by choosing the largest face similarity across all levels.
\cref{fig:layerwise} (last column) shows the results from the first two strategies.
Obviously, \method from the 5-th to the 9-th levels best preserve the identity information.
\cref{tab:comparison}b compares \method with other unsupervised feature learning methods, including VAE~\cite{vae}, MoCo~\cite{moco}, and ALAE~\cite{alae}.
All these competitors are also trained on FF-HQ dataset~\cite{stylegan} with optimally chosen hyper-parameters.
ResNet-50~\cite{resnet} is employed as the backbone for MoCo and \method.
Our method with voting strategy achieves 69.7\% accuracy, surpassing other competitors by a large margin.
We also visualize some reconstructed LFW faces in \cref{fig:lfw}, where our \method well handles the domain gap (\textit{e.g.}, image resolution) and preserves the identity information.

\subsubsection{Large-Scale Image Classification}\label{sec:imagenet}
We further evaluate \method on the high-level image classification task using ImageNet~\cite{imagenet}.
Before the training of encoder, we first train a StyleGAN model, with $256\times256$ resolution, on the ImageNet training collection.
After that, we learn the hierarchical encoder by using the pre-trained generator as the supervision.
No labels are involved in the above training process.%
\footnote{Our encoder can be trained very efficiently, usually $3\times$ faster than the GAN training.}
For the image classification problem, we train a linear model on top of the features extracted from the training set with the softmax loss.
Then, this linear model is evaluated on the validation set.%
\footnote{During testing, we adopt the fully convolutional form as in \cite{googlenet} and average the scores at multiple scales.}
\cref{tab:imagenet_cls} shows the comparison between \method and other unsupervised representation learning approaches~\cite{instdisc,cpc,moco,bigan,ssgan,bigbigan}, where we beat most of the competitors.
The state-of-the-art MoCo~\cite{moco} gives the most compelling performance.
But different from the representations learned with contrastive learning, \method has huge advantages in generative tasks, as already discussed in \cref{exp:generative}.
Among adversarial representation learning approaches, BigBiGAN~\cite{bigbigan} achieves the best performance, benefiting from the incredible large-scale training.
\revise{However, \method presents a stronger ability for image reconstruction. 
BigBiGAN is learned by discriminating the data-latent joint distribution, while our \method targets image reconstruction by treating a well-trained GAN generator as a learned loss function. 
Consequently, as shown in Fig.~15, BigBiGAN can only recover the input images from the category level, instead, our approach can recover the inputs with much more details. 
The reconstruction error in Tab.~8 conveys the same conclusion. 
This is also the reason why \method could facilitate various low-level and middle-level discriminative tasks beyond image classification.
More details about ImageNet training can be found in \textbf{Supplementary Material}.}

\setlength{\tabcolsep}{8pt}
\begin{table}[t]
  \caption{
    Quantitative comparison on the ImageNet~\cite{imagenet} classification task.
  }
  \vspace{-10pt}
  \label{tab:imagenet_cls}
  \centering
  \begin{tabular}{llc}
    \toprule
    Method                                   & Architecture & Top-1 Acc. \\ \midrule
    Motion Seg (MS)~\cite{motionseg,carl}    &   ResNet-101 &       27.6 \\
    Exemplar (Ex)~\cite{exemplar,carl}       &   ResNet-101 &       31.5 \\
    Relative Po (RP)~\cite{relativepos,carl} &   ResNet-101 &       36.2 \\
    Colorization (Col)~\cite{colorful,carl}  &   ResNet-101 &       39.6 \\ \midrule
    \multicolumn{3}{c}{\textit{Contrastive Learning}}                    \\[2pt]
    InstDisc~\cite{instdisc}                 &    ResNet-50 &       42.5 \\
    CPC~\cite{cpc}                           &   ResNet-101 &       48.7 \\
    MoCo~\cite{moco}                         &    ResNet-50 &       60.6 \\ \midrule
    \multicolumn{3}{c}{\textit{Generative Modeling}}                     \\[2pt]
    BiGAN~\cite{bigan}                       &      AlexNet &       31.0 \\
    SS-GAN~\cite{ssgan}                      &    ResNet-19 &       38.3 \\
    BigBiGAN~\cite{bigbigan}                 &    ResNet-50 &       55.4 \\ \midrule 
    \method (Ours)                           &    ResNet-50 &        51.1 \\ \bottomrule
  \end{tabular}
  \vspace{-10pt}
\end{table}

\setlength{\tabcolsep}{13pt}
\begin{table}[t]
  \caption{
    Qualitative comparison between BigBiGAN~\cite{bigbigan} and \method on reconstructing images from ImageNet~\cite{imagenet}.
  }
  \label{tab:image_rec}
  \vspace{-10pt}
  \centering\small
  \begin{tabular}{lccc}
    \toprule
                             &  \MSE & \SSIM &  \FID \\ \midrule
    BigBiGAN~\cite{bigbigan} & 0.363 & 0.236 & 33.42 \\
    \method (Ours)           & 0.078 & 0.431 & 22.70 \\ \bottomrule
  \end{tabular}
  \vspace{-5pt}
\end{table}

\begin{figure*}[t]
  \centering
  \includegraphics[width=1.0\linewidth]{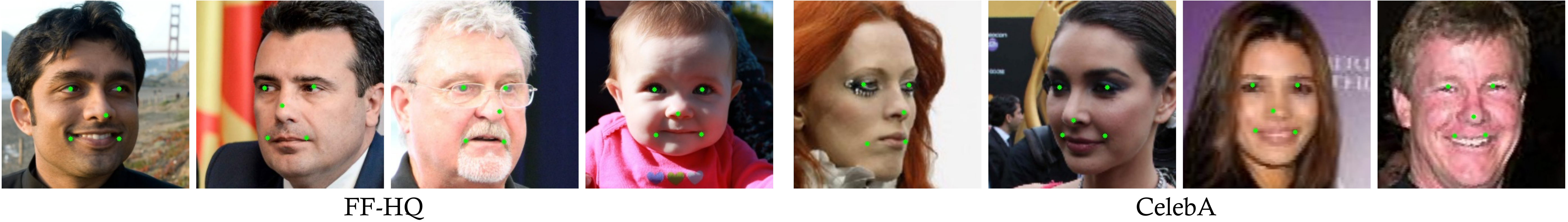}
  \vspace{-22pt}
  \caption{
    \textbf{Landmark detection} results.
    \method is trained on FF-HQ~\cite{stylegan} dataset but can successfully handle the hard cases (large pose and low image quality) in MAFL dataset~\cite{tcdcn}, a subset of CelebA~\cite{celeba}.
  }
  \label{fig:landmark}
  \vspace{-7pt}
\end{figure*}

\begin{figure*}[t]
  \centering
  \includegraphics[width=1.0\linewidth]{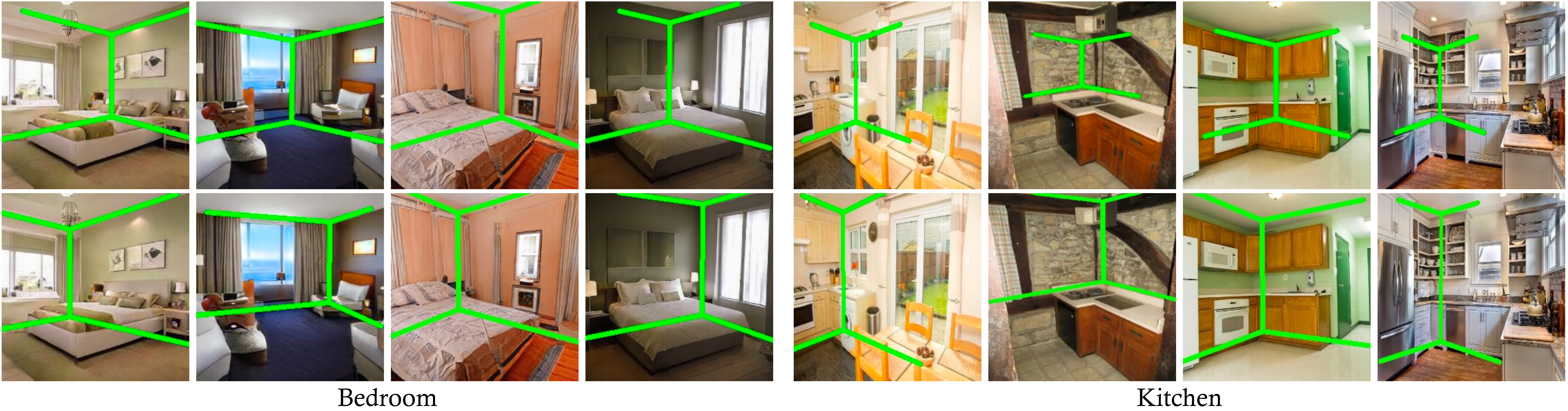}
  \vspace{-22pt}
  \caption{
    \textbf{Layout prediction} results using feature learned by MoCo~\cite{moco} (top row) and our \method (bottom row).
    Both methods are trained on LSUN bedrooms~\cite{lsun} and then transferred to LSUN kitchens.
  }
  \label{fig:layout}
  \vspace{-10pt}
\end{figure*}

\setlength{\tabcolsep}{15pt}
\begin{table}[t]
  \caption{Landmark detection results on MAFL~\cite{tcdcn}. \revise{\method-R denotes \method trained with regularizer.}}
  \label{tab:landmark}
  \vspace{-10pt}
  \centering\small
  \begin{tabular}{lcc}
    \toprule
    Method                                   & Supervision &  \MSE \\ \midrule
    TCDCN~\cite{tcdcn}                       &   \ding{51} &  7.95 \\
    MTCNN~\cite{mtcnn}                       &   \ding{51} &  5.39 \\
    Cond. ImGen~\cite{jakab2018unsupervised} &             &  4.95 \\
    ALAE~\cite{alae}.                        &             & 10.13 \\
    MoCo-R50~\cite{moco}                     &             &  9.07 \\ 
    \revise{CLIP-R50~}                       &             &  \revise{4.98} \\
    \midrule
    \method (Ours)                           &             &  5.12 \\
    \revise{\method-R}                  &             &  \revise{4.92}  \\
    \bottomrule
  \end{tabular}
  \vspace{-10pt}
\end{table}

\subsubsection{Transfer Learning}\label{exp:transfer}
In this part, we explore how \method can be transferred from one dataset to another.

\noindent\textbf{Landmark Detection.}
We train a linear regression model using \method on FF-HQ~\cite{stylegan} and test it on MAFL~\cite{tcdcn}, which is a subset of CelebA~\cite{celeba}.
This two datasets have a large domain gap, \textit{e.g.}, faces in MAFL have larger poses yet lower image quality.
As shown in \cref{fig:landmark}, \method shows a strong transferability across these two datasets.
We compare our approach with some supervised and unsupervised alternatives~\cite{tcdcn,mtcnn,jakab2018unsupervised,moco}.
\revise{CLIP~\cite{radford2021clip} trained with 400,000,000 image-text paired samples is also included to serve as a strong baseline to compare with \method. }
For a fair comparison, we try the multi-scale representations from MoCo~\cite{moco} and \revise{CLIP~\cite{radford2021clip}} (\textit{i.e.}, Res2, Res3, Res4, and Res5 feature maps) and report the best results.
\cref{tab:landmark} demonstrates the strong generalization ability of \method.
In particular, it achieves on-par or better performance than the methods that are particular designed for this task~\cite{tcdcn,mtcnn,jakab2018unsupervised}.
Also, it outperforms MoCo~\cite{moco} on this mid-level discriminative task.
\revise{As the \cref{tab:landmark} below suggests, \method achieves comparable performance as CLIP-R50 with significantly better data efficiency.
Such a comparison is not 100\% eye-to-eye because our approach is particularly trained on human faces while CLIP could cover a much larger data domain.
But it still demonstrates, to some extent, that adequately leveraging the pre-trained GAN generator as a learned loss function yields a discriminative and transferable visual representation.}

\noindent\textbf{Layout Prediction.}
We train the layout predictor on LSUN~\cite{lsun} bedrooms and test it on kitchens to validate how \method can be transferred from one scene category to another.
Feature learned by MoCo~\cite{moco} on the bedroom dataset is used for comparison.
We can tell from \cref{fig:layout} that \method shows better predictions than MoCo, especially on the target set (\textit{i.e.}, kitchens), suggesting a stronger transferability.
Like landmark detection, we also conduct experiments with the 4-level representations from MoCo~\cite{moco} and select the best.

\begin{figure}[t]
  \centering
  \includegraphics[width=1.0\linewidth]{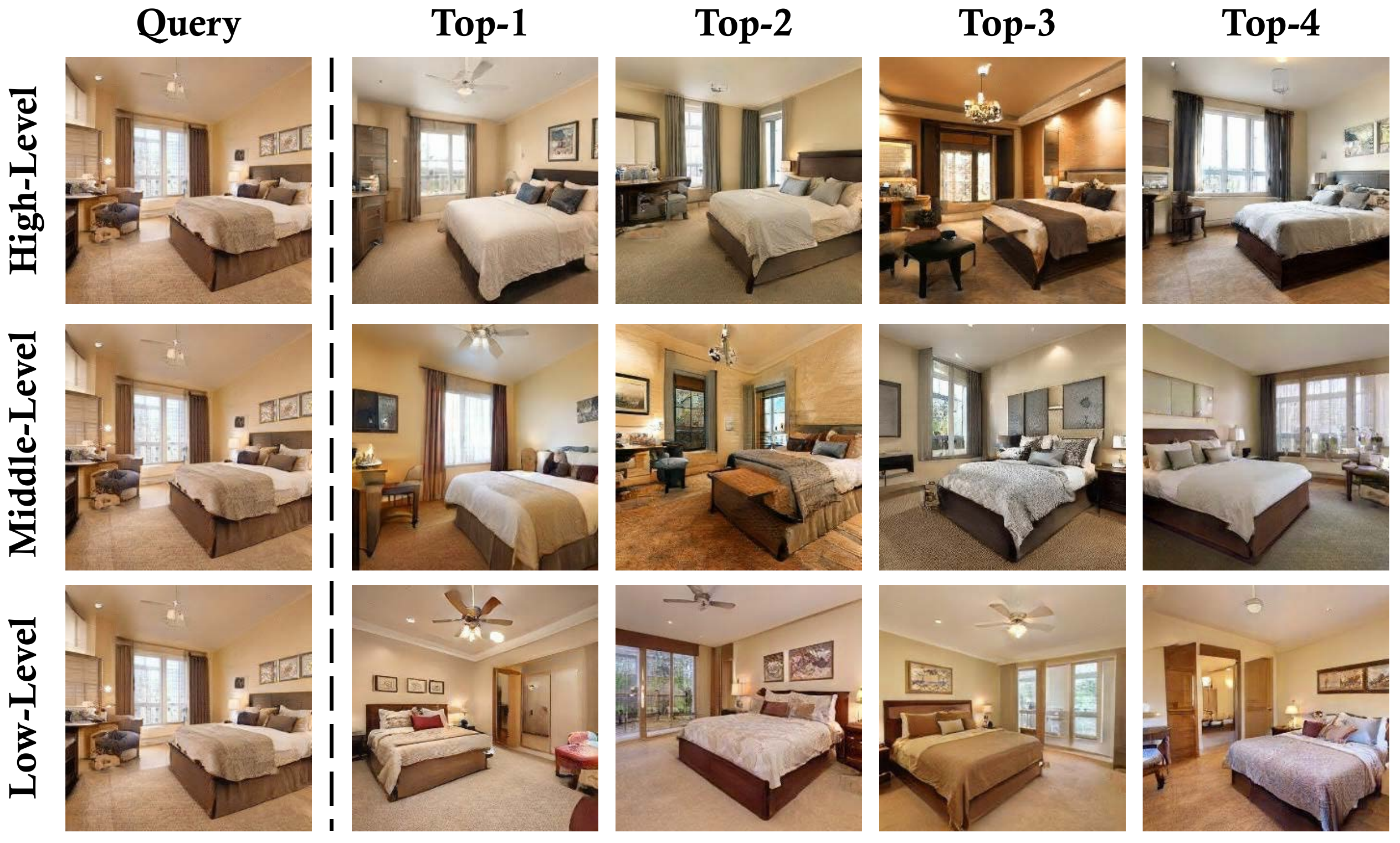}
  \vspace{-24pt}
  \caption{
    Retrieval results on LSUN bedroom~\cite{lsun}.
    %
  }
  \label{supp:fig:re_bedroom}
  \vspace{-7pt}
\end{figure}

\begin{figure*}[t]
  \centering
  \includegraphics[width=1.0\linewidth]{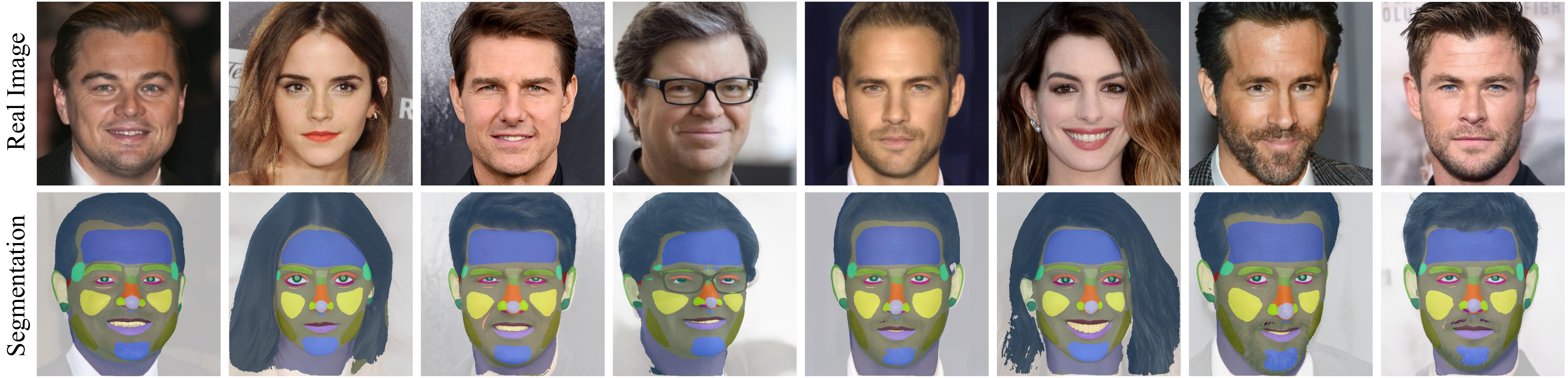}
  \vspace{-20pt}
  \caption{
    \textbf{Data-efficient Image Segmentation} with Spatial \method.
    We use the spatial-aware encoder to obtain a set of generative features with spatial dimension and a segmentation head trained with limited annotated data to obtain segmentation results.
    }
  \label{fig:segmentation}
  \vspace{-5pt}
\end{figure*}

\begin{figure*}[t]
  \centering
  \includegraphics[width=1.0\linewidth]{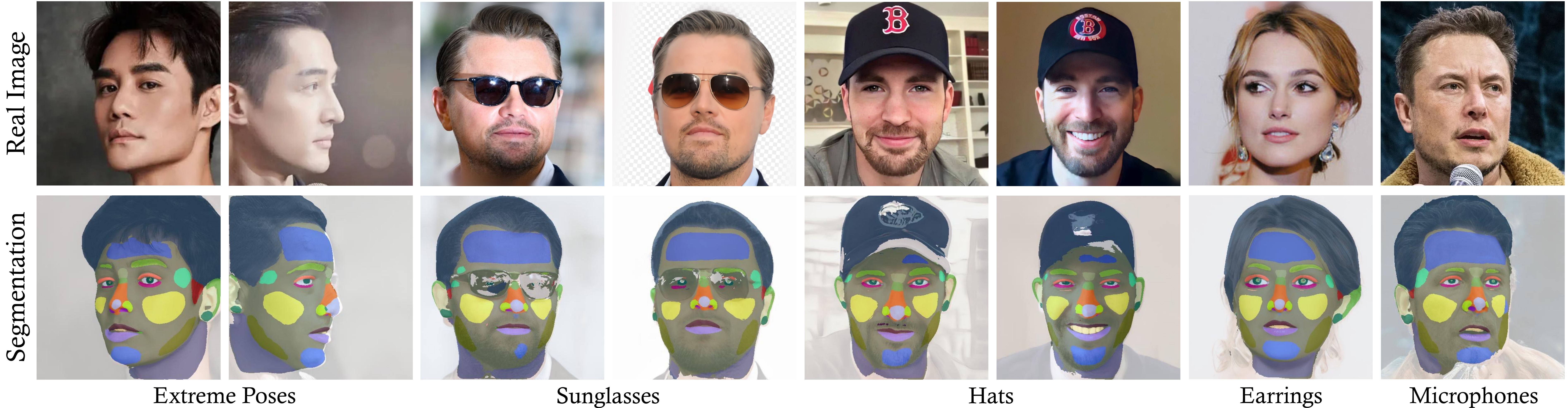}
  \vspace{-20pt}
  \caption{
    \revise{Extreme cases of data-efficient image segmentation with Spatial \method.}
    These extreme samples (\textit{i.e.} extreme pose, hat, sunglasses, earrings as well as microphones) show the robustness of the segmentation head only trained with fewer annotated samples.
    }
  \label{fig:segmentation-excases}
  \vspace{-5pt}
\end{figure*}

\subsubsection{Image Retrieval}\label{exp:retrieval}
In this section, we verify the hierarchical property of the proposed \method with image retrieval.
Concretely, given a query image, we use encoder to extract its \method.
Then, we use different levels of \method to perform retrieval from $10K$ real images.
Note that \method from these $10K$ images are prepared in advance and $\ell_1$ distance is used as the metric for retrieval.
\cref{supp:fig:re_bedroom} shows the retrieval results on LSUN bedroom~\cite{lsun}.
We can tell that when we use higher level (first row) features for retrieval, all retrieved results are with the same layout as the query image, but they may have different lighting conditions.
Meanwhile, when using lower level (bottom row) features for retrieval, the retrieved results are with similar lighting condition as the query image.

\subsection{Spatial Expansion}\label{exp:spatial-task}

\subsubsection{Spatial \method} 

\noindent{\textbf{Spatial-Aware Style Codes.}} Even though the layer-wise style codes can describe the global semantics of synthesized images, the fine-grained semantics cannot be expressed precisely because the style codes are too coarse to maintain spatial semantics. 
To facilitate the style codes with semantic segmentation, we equip the layer-wise style codes with spatial dimension. 
It is noteworthy that the introduced spatial dimension make the layer-wise representation more flexible for various of vision tasks.

\noindent{\textbf{Spatial-Aware Encoder}}. 
For the vision tasks requiring the spatial-aware representation of the input image, a spatial-aware encoder is also needed to produce the spatial-aware style codes.
We inherit the backbone and FPN to fuse the semantics encoded at different level.
The last three stages feature maps $\{R_4, R_5, R_6\}$, are used to produce spatial-aware \method.
We also use the same instantiation for the layer equipment.
But differently, we use an $1\times1$ convolution layer to embed the feature maps  $\{R_4, R_5, R_6\}$ and an upsampler to match the spatial size of the corresponding convolution feature map. 
It can be formulated as: 
\begin{align}
  GH_j &= \mathtt{up}(W_j R_{a[j]}, h_{C_j}/h_{R_{a[j]}}) \quad j \in \{1, N\},\nonumber
\end{align}
where  $GH_j$ is the learned spatial-aware representation, $C_j$ is the convolutional feature map,  $a[j]$ denotes the corresponding index of the output feature map from FPN, and $h_{C_j}, h_{R_{a[j]}}$ denotes the spatial dimension of feature map $C_j$ and $R_{a[j]}$.

\begin{figure}[t]
  \centering
  \includegraphics[width=1.0\linewidth]{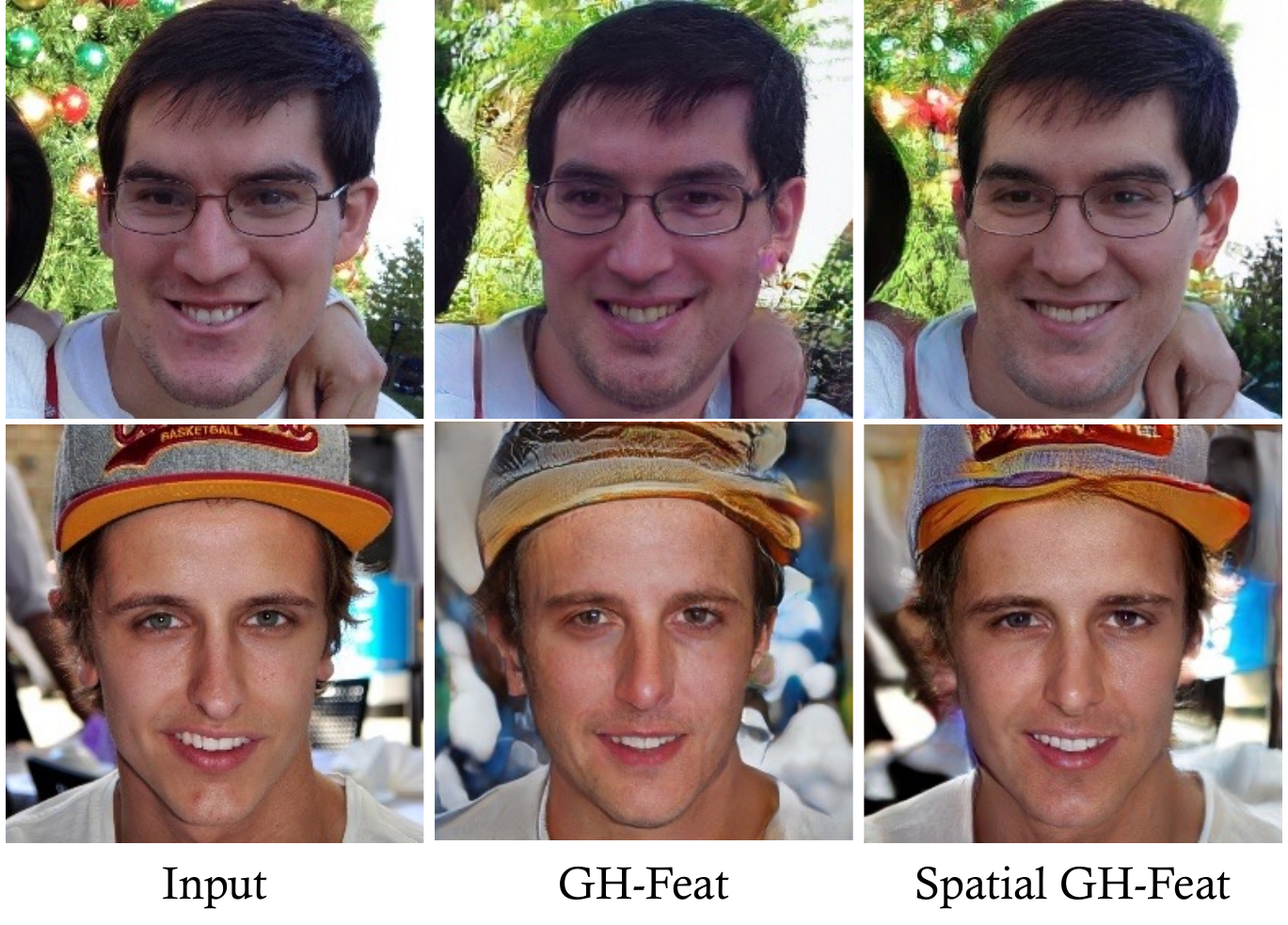}
  \vspace{-20pt}
  \caption{
    Qualitative comparison between \method and spatial \method.
  }
  \label{fig:spatial-ablation}
  \vspace{-5pt}
\end{figure}

\noindent{\textbf{Ablation.}} 
The proposed spatial generative feature is adopted to provide spatial information, and thus it is critical to the quality of the reconstructed image.
As shown in \cref{tab:spatial-inversion}, the spatial generative feature can improve the reconstruction performance,
and the qualitative results in \cref{fig:spatial-ablation} present that the spatial \method is able to reconstruct the background and the out-of-the-distribution objects \textit{i.e.} hands and hats well.
It supports the effectiveness of the spatial-aware generative features.

\subsubsection{Data-Efficient Semantic Segmentation}
Compared with classification, image segmentation needs more precise prediction along the spatial dimension.
However, the generative features without spatial dimension cannot facilitate this task because they cannot be aware of the semantics for each pixel.
To enable this task, we use the spatial-aware encoder to obtain a set of generative features with spatial dimension, and a segmentation head $\textit{i.e.}$ the Style Interpreter in \cite{datasetgan} is followed to obtain the segmentation results.   
Because of the generalization of the spatial visual features, we only need a few samples to achieve a good segmentation head. 
In our experiment, we used 20 annotated samples for the training.
We visualize predictions learned from our visual features in \cref{fig:segmentation}.
Obviously, the spatial-aware generative features provide precise information for dense pixels, facilitating image segmentation with a few annotations.


\revise{We include several extreme cases in Fig. \ref{fig:segmentation-excases} to verify the robustness of the segmentation results achieved by \method.
Concretely, we include samples under extreme poses, as well as samples containing out-of-distribution objects (\textit{i.e.}, the objects without annotations during the training of the segmentation branch).
We have three observations:
(1) Even there are few samples under extreme poses during training, our approach could still produce promising segmentation results on such challenging cases at the inference stage.
(2) The model could well recognize the eyeglass frames yet perform poorly on eyeglass lens. We guess this is caused by the overlap between lens and eyes.
(3) Hats (recognized as hair), earrings and microphones (recognized as background) could be regarded as failure cases, because our segmentation branch is learned with simple annotations (\textit{e.g.}, eyes, nose, cheek, \textit{etc.}). A more competitive performance could be expected given richer segmentation labels.}

\setlength{\tabcolsep}{7.5pt}
\begin{table}[t]
  \caption{
    Quantitative comparison on image reconstruction between \method and spatial \method.
    \revise{\method-R denotes \method trained with regularizer.}
  }
  \label{tab:spatial-inversion}
  \vspace{-10pt}
  \centering\small
  \begin{tabular}{lcccc}
    \toprule
            & \multicolumn{2}{c}{Face} & \multicolumn{2}{c}{Bedroom} \\ \cmidrule[0.5pt](lr){2-3} \cmidrule[0.5pt](lr){4-5}
    Method           &  \MSE & \SSIM &  \MSE & \SSIM  \\ \midrule
    \method   & 0.046 & 0.56  & 0.068 & 0.52  \\
    \revise{\method-R}  & \revise{0.049} & \revise{0.55} & \revise{0.070} & \revise{0.50} \\
    Spatial \method   & 0.029 & 0.67 & 0.057 & 0.58 \\
    \bottomrule 
  \end{tabular}
  \vspace{-5pt}
\end{table}

\vspace{-1pt}
\section{Conclusion}\label{sec:conclusion}
\vspace{-2pt}
In this work, we consider the well-trained GAN generator as a learned loss function for learning multi-scale features. 
Unlike previous work, we treat layer-wise style codes in $\Y$ space as generative visual features rather than $\W$ space, resulting in better hierarchical properties.
A distribution-level regularizer is introduced to overcome the limitation of only using image-level supervision for encoder training.
The resulting Generative Hierarchical Features are shown to be generalizable to a wide range of vision tasks.
Since GH-Feat only leverages the semantics learned in GANs, the features may lack the good properties of the discriminative model features. 
In the future, we hope to learn deep representations by unifying discriminative and generative models that can complement each other.
%

\ifCLASSOPTIONcompsoc
  \section*{Acknowledgments}
\else
  \section*{Acknowledgment}
\fi
This work is supported in part by the Early Career Scheme (ECS) through the Research Grants Council (RGC) of Hong Kong under Grant No.24206219, Grant No.14204521, CUHK FoE RSFS Grant, and Centre for Perceptual and Interactive Intelligence (CPII) Ltd under the Innovation and Technology Fund.

\ifCLASSOPTIONcaptionsoff
  \newpage
\fi
\bibliographystyle{IEEEtran}
\bibliography{references}

\newpage
\begin{figure*}[t]
  \centering
  \includegraphics[width=0.95\linewidth]{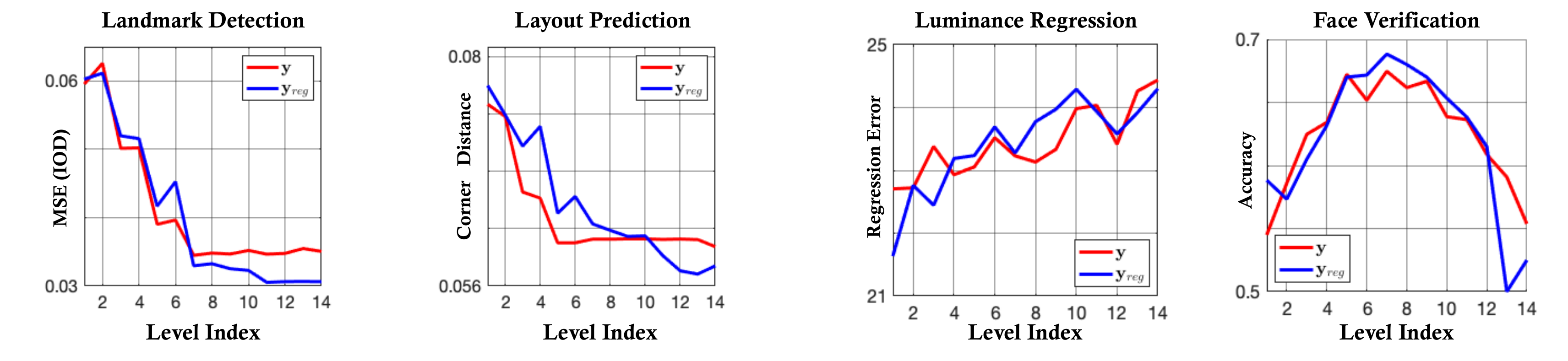}
  \vspace{-20pt}
  \caption{
    Comparison on the hierarchical property between using regularizer or not. 
    $\y$ (in \textbf{\textcolor{red}{red}} color) and $\y_{reg}$ (in \textbf{\textcolor{blue}{blue}} color) denote the original \method and \method with regularizer, respectively.
    %
    }
  \label{fig:layerwise-reg}
  \vspace{-5pt}
\end{figure*}

\section*{A1. Hierarchical Property}

We also re-evaluate the layer-wise representation on different discriminative tasks.
As shown in Fig. \ref{fig:layerwise-reg}, the training regularizer improves the hierarchical property of the original \method.
Since the training regularizer prevents the model from overfitting pixel values, the layer-wise representation is closer to the distribution center and achieve better hierarchical properties on the discriminative tasks. %

\section*{A2. Experiments on ImageNet}\label{supp:sec:imagenet}

\begin{figure*}[!ht]
  \centering
  \includegraphics[width=0.95\linewidth]{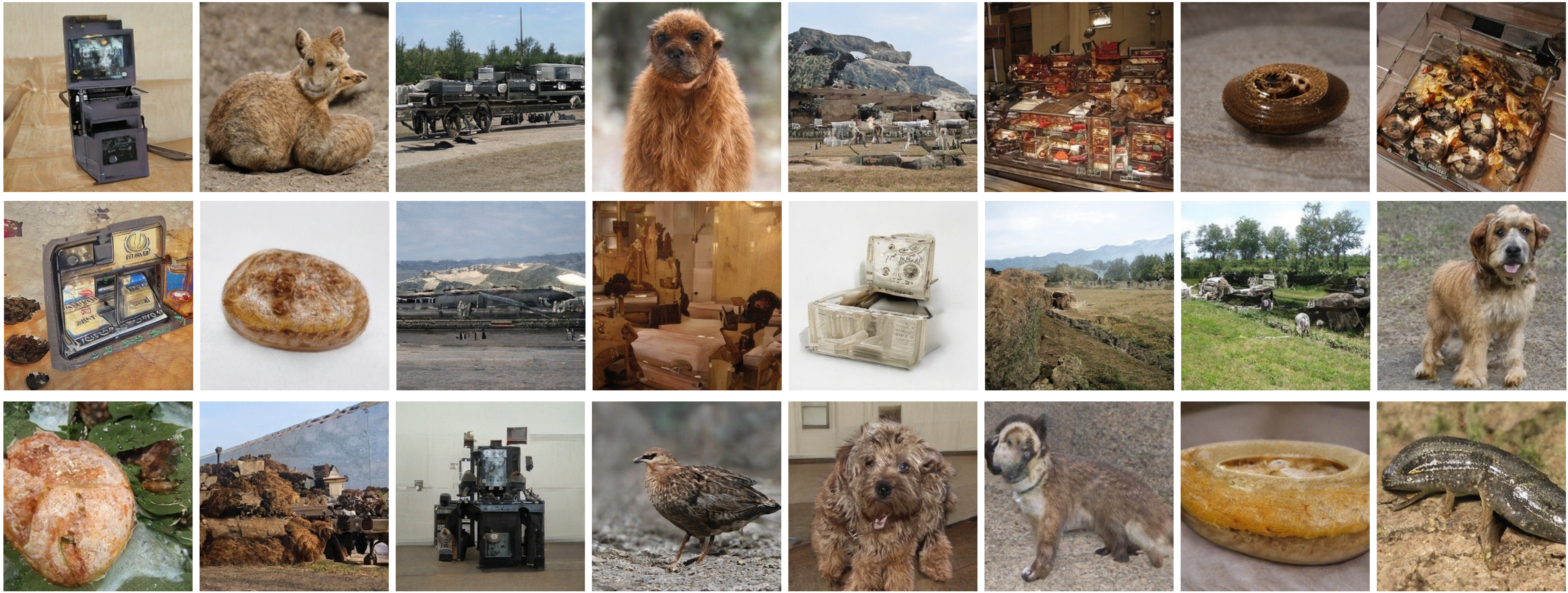}
  \caption{
    \textbf{Uncurated generated samples} of StyleGAN model on ImageNet.
    }
  \label{fig:imagenet-samples}
\end{figure*}

\vspace{2pt}
\noindent{\textbf{Training Details.}}
During the training of the StyleGAN model on the ImageNet dataset~\cite{imagenet}, we resize all images in the training set such that the short side of each image is 256, and then centrally crop them to $256\times256$ resolution.
All training settings follow the StyleGAN official implementation~\cite{stylegan_github}, including the progressive strategy, optimizer, learning rate, \textit{etc}.
The generator and the discriminator are alternatively optimized until the discriminator have seen $250M$ real images.
After that, the generator is fixed and treated as a well-learned loss function to guide the training of the encoder.
During the training of the hierarchical encoder, images in the training collection are pre-processed in the same way as mentioned above.
After the encoder is ready, we treat it as a feature extractor.
%
We use the output feature map at the ``res$_5$'' stage, apply adaptively average pooling to obtain $2\times2$ spatial feature and vectorize it.
A linear classifier, \textit{i.e.}, with one fully-connected layer, takes these extracted features as the inputs to learn the image classification task.
SGD optimizer, together with batch size 2048, is used.
The learning rate is initially set as 1 and decayed to 0.1 and 0.01 at the 60-th and the 80-th epoch respectively.
During the training of the final classifier, ResNet-style data augmentation~\cite{resnet} is applied.

The FID score on ImageNet is 40.92.
Fig~\ref{fig:imagenet-samples} shows the uncurated samples of the pretrained ImageNet samples.
Although the synthesized samples are not very realistic, they can still help downstream tasks like ImageNet classification. 

\vspace{2pt}
\noindent{\textbf{Discussion.}}
We have already shown in the main submission that \method achieves comparable accuracy to existing alternatives.
Especially, among all methods based on generative modeling, \method obtains second performance only to BigBiGAN~\cite{bigbigan}, which requires incredible large-scale training.
However, as discussed in the main submission, our \method facilitates a wide rage of tasks besides image classification.
Taking image reconstruction as an example, our approach can well recover the input image, significantly outperforming BigBiGAN~\cite{bigbigan}.

\end{document}